\documentclass{article}

\usepackage{graphicx}
\usepackage{algorithm}
\usepackage{algpseudocode}
\usepackage{mathtools}
\usepackage[symbol]{footmisc}
\usepackage{subcaption}
\newtheorem{definition}{Definition}
\usepackage{amssymb}
\usepackage{placeins}
\usepackage{multirow}
\usepackage{hyperref}
\usepackage{booktabs}
\usepackage{siunitx}
\usepackage{tabularx}

\usepackage{geometry}
\usepackage{amsmath}

\title{Going with the Flow: Approximating Banzhaf Values via Graph Neural Networks}
\author{
  \begin{tabular}{c}
    \textbf{Benjamin Kempinski} \\
    \texttt{benjamin.kempinski@donders.ru.nl}
  \end{tabular}
  \qquad
  \begin{tabular}{c}
    \textbf{Tal Kachman} \\
    \texttt{tal.kachman@ru.nl}
  \end{tabular} \\[0.5em]
  Radboud University\\Donders Institute and Department of AI\\Nijmegen, Netherlands
}
\date{}
\begin{document}
\maketitle


\begin{abstract}
Computing the Banzhaf value in network flow games is fundamental for quantifying agent influence in multi-agent systems, with applications ranging from cybersecurity to infrastructure planning. However, exact computation is intractable for systems with more than $\sim20$ agents due to exponential complexity $\mathcal{O}(2^m)$. While Monte Carlo sampling methods provide statistical estimates, they suffer from high sample complexity and cannot transfer knowledge across different network configurations, making them impractical for large-scale or dynamic systems.
We present a novel learning-based approach using Graph Neural Networks (GNNs) to approximate Banzhaf values in cardinal network flow games. By framing the problem as a graph-level prediction task, our method learns generalisable patterns of agent influence directly from network topology and control structure. We conduct a comprehensive empirical study comparing three state-of-the-art GNN architectures—Graph Attention Networks (GAT), Graph Isomorphism Networks with Edge features (GINE), and EdgeConv—on a large-scale synthetic dataset of 200,000 graphs per configuration, varying in size (20-100 nodes), agent count (5-20), and edge probability (0.5-1.0).
Our results demonstrate that trained GNN models achieve high-fidelity Banzhaf value approximation with order-of-magnitude speedups compared to exact and sampling-based methods. Most significantly, we show strong zero-shot generalisation: models trained on graphs of a specific size and topology accurately predict Banzhaf values for entirely new networks with different structural properties, without requiring retraining. This work establishes GNNs as a practical tool for scalable cooperative game-theoretic analysis of complex networked systems.
\end{abstract}



\section{Introduction}
Network flow models are fundamental to modelling and optimising real-life complex systems. For decades, they have enabled applications from logistics to energy grids ~\cite{granot_1992, Ford_Fulkerson_1956, graph_data_mining}. Classical flow problems focus on finding the optimal paths. A critical layer of complexity arises when network components, such as edges or nodes, are owned and controlled by autonomous, self-interested agents. In these multi-agent settings, the overall network capacity is not a static property, but rather an emergent one that depends on strategic interactions and cooperation of these agents\cite{10.1007_s10458-025-09696-7,flow_coalition_2010}. Understanding and quantifying the influence of each agent over the system's total throughput is therefore a central challenge in designing, analysing and securing such systems. 

Cooperative game theory provides a principled framework for analysing these agent-driven networks. Game theory has long been used to assess centrality, connectiveness and flow \cite {Koschtzki2005, 10.5555/1062400, DUNNE201020, NAMBUI20181012, Huang2024}. Modelling the system as a Cardinal Network Flow Game~\cite{10.1007/s10458-008-9057-6}. We can formally characterise the value generated by any "coalition" of cooperating agents as the maximum flow they can collectively enable ~\cite{Bachrach2010-rw}. Within this framework, a key objective is to assign a measure of power or importance to each agent. The Banzhaf value is a canonical solution concept for quantifying an agent's power as their average marginal contribution to the network's capacity across all possible coalitions~\cite{bachrach2007computing}. This measure is crucial for resource allocation and identifying critical agents~\cite {6139974, LEVY2011411}

Despite its desirable theoretical properties, practical applications of the Banzhaf value are severely hampered by its computational complexity\cite{jiralerspong2024expectedflownetworksstochastic}. The exact calculation requires iterating through all possible coalitions, resulting in a complexity of $\mathcal{O}(2^m)$ for a game with $m$ agents \cite{415dc80a-289c-39e1-8e6b-601fc5ef267e}. Combined with maximum flow computation complexity of $\mathcal{O}(E* V^2)$ \cite{10-1145-321694-321699}, this renders exact computation infeasible beyond $\sim20$ agents (which would require more than   $2^{20}=1,048,576$ coalitions). These limitations preclude the analysis of many real-world problems of significant interest, such as the stability of complex biological and chemical networks\cite{Cesari2018-qf,thöni2025modeling}, large-scale cybersecurity scenarios \cite{10.1007/978-3-030-01554-1_20}, and military and defensive planning \cite{e3fb9f6d7c2d41478142c43ff6272b33,9699407,drones2006}, which routinely have hundreds and thousands of interacting components.

Existing approximation methods, like Monte Carlo sampling\cite{Bachrach2010-rw,bachrach2007computing} and Marginal-Contribution Networks\cite{mcnet_shapley_2009} methods, provide statistical estimates of the Banzhaf value~\cite{Bachrach2010-rw} but suffer from high sample complexity: Monte Carlo approaches require $\mathcal{O}(n^2/\epsilon^2)$ coalition evaluations for $\epsilon$-approximation, while Marginal Contribution Networks need polynomial samples in network size. Each evaluation requires solving a maximum flow problem, making convergence slow even for moderate-sized networks. Moreover, these methods are stateless, requiring recomputation for any network change. This inability to generalise motivates a learning-based approach that can efficiently approximate the Banzhaf value while learning generalisable principles of agent influence. Learning-based methods have shown promise in simpler game settings\cite{kempinski2025,diazortiz2023usingcooperativegametheory,cornelisse2022neuralpayoffmachinespredicting}. We propose using Graph Neural Networks (GNNs), deep learning models designed to learn from relational graph structure\cite{wang2019dynamicgraphcnnlearning,gilmer2017neuralmessagepassingquantum,xu2019powerfulgraphneuralnetworks,veličković2018graphattentionnetworks}.

We showcase how GNNs provide a fast, accurate, and generalisable method for approximating Banzhaf values in network flow games. We introduce a novel framework by framing the problem as a graph-level prediction task, where network topology and agent control structure are encoded as input features. We conduct a large-scale empirical study with comprehensive synthetic datasets, enabling rigorous comparison of state-of-the-art GNN architectures. We demonstrate that trained GNN models approximate Banzhaf values with high fidelity, achieving order-of-magnitude speedups compared to exact and sampling-based methods. 
Most critically, we show strong zero-shot generalisation: models accurately predict Banzhaf values for entirely new graphs with different sizes and topologies than those seen during training. 

\begin{figure*}[t] 
     \centering
     \includegraphics[width=0.9\textwidth]{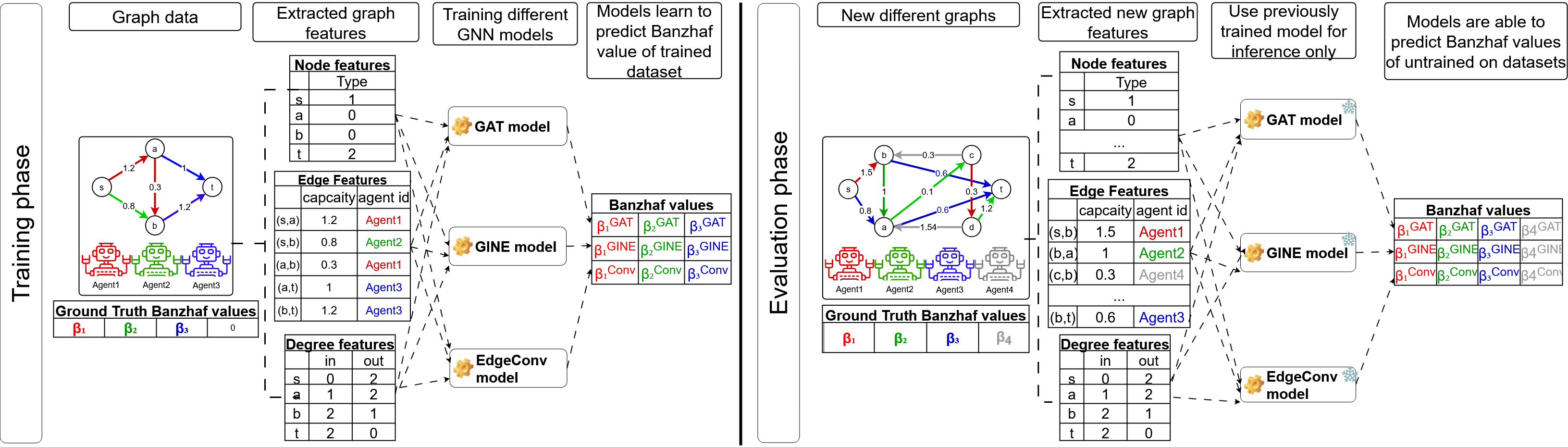}
     \caption{Overview of our GNN-based approach for Banzhaf value prediction. Left (Training phase): A network flow graph with edges controlled by three agents (colour-coded). Node features(type indices), edge features(normalised capacity and agent control) and degree features(in/out degree) are extracted and processed by three GNN architectures (GAT, GINE, EdgeConv) to predict the Banzhaf value vector. Right (Evaluation phase): The same trained models are applied to entirely different graph structures with more nodes and agents, demonstrating zero-shot generalisation across network topologies without additional training.}
     \label{fig:cartoon}
\end{figure*}
   
\section{Related Work}
Our research is at the interface of cooperative game theory and graph representation learning. We examine our contribution by reviewing the evolution of methods of power index approximation and the application of machine learning to graph-structured problems.

The challenge of computing power indices is fundamental. Both Banzhaf value and the closely related Shapley value\cite{Banzhaf_1965_5380,Shapley_Shubik_1954} are \#P-hard to compute in the general case~\cite{Deng1994OnTC,bachrach2016analyzing,michalak2013efficient}. This complexity class signifies that an efficient, exact algorithm is highly unlikely to exist, necessitating the use of approximation methods for any non-trivial problem size.

While both measure player importance, they differ axiomatically. The Shapley value satisfies efficiency (the values sum to the total game value), symmetry, null player, and additivity (linearity over game combinations) axioms. The Banzhaf value satisfies the symmetry, null player and additivity, but does not satisfy efficiency, instead focusing on a player's direct marginal contribution. In many domains, including feature attribution and network analysis, the Banzhaf value's intuitive interpretation of marginal impact and its slightly lower computational cost have made it a compelling alternative to the Shapley value~\cite{pmlr-v180-karczmarz22a}. Given this context, the development of scalable approximation techniques has been a primary focus of research for decades.

The field of explainable AI has driven a surge of innovation in power index approximation, where the Shapley and Banzhaf values are used to attribute a model's prediction to its input features. The landmark SHAP framework introduced kernel-based methods and model-specific optimisations to make Shapley values a practical tool for interpreting machine learning models\cite{lundberg2017unifiedapproachinterpretingmodel}. This success has inspired exploration of learning-based approaches that can amortise computational cost across instances. 

The predominant approach for estimating power indices (and similar notions) is Monte-Carlo simulation, which approximates the average marginal contribution by sampling random coalitions \cite{Bachrach2010-rw,castro2009polynomial,bachrach2016misrepresentation}. Unbiased sampling methods can converge slowly and require many samples to achieve low-variance estimates for a larger number of agents. Refinement, such as stratified and quasi-Monte Carlo sampling, has been proposed to improve sample efficiency, but the core limitation remains: the approximation is performed on a per-instance basis~\cite{shapley_survey_2025}.
\footnote{Similarly, multiple samples must be taken in stochastic settings to overcome the variance triggered by the stochastic process~\cite{zuckerman2008manipulating,bachrach2012solving}.}

This has inspired a new paradigm of surrogate modelling for value approximation. Methods like FastSHAP~\cite{jethani2022fastshaprealtimeshapleyvalue} and \hspace{0pt}KernelBanzhaf~\cite{liu2025kernelbanzhaffastrobust} train a separate machine learning model to predict the output of a game-theoretic function. Our work, while similar in its motivation, is different in approach. 

FastSHAP trains on noisy Monte-Carlo estimates, trading accuracy and repeatability for speed. KernelBanzhaf uses partial Monte-Carlo sampling with linear regression to balance accuracy and runtime. Both approaches estimate the power indices values using some stochastic approximation, and then train a model to go over the sampled data and learn to estimate the correct value in reduced time and with good comparability to the Monte-Carlo method. However, they can only be applied to the system for which they were trained. Systems of different features or agent spaces require renewed training and fitting. Our work presents the use of Graph Neural Networks(GNN), a class of deep learning models designed to learn functions on graph-structured data by iteratively passing and aggregating information between nodes\cite{kipf2017semisupervisedclassificationgraphconvolutional, veličković2018graphattentionnetworks, xu2019powerfulgraphneuralnetworks}. The key strength of GNNs is in their ability to learn patterns that are invariant to permutations of the node ordering and applicable to graphs of varying sizes. While GNNs have been used to approximate structural graph metrics like node-centrality~\cite{LI2024111174}, their application to learning cooperative game-theoretic values—which depend on the exponential and non-local evaluation of subsets—remains largely unexplored. Our work aims to use GNNs not just to learn a structural property of a graph, but to approximate the outcome of a complex, global cooperative game played upon it, and to generalise this knowledge to entirely new game instances.

\section{Notations and preliminary}
This section establishes the theoretical foundations required for our analysis, drawing from cooperative game theory and network flow theory \cite{Bachrach2010-rw,384b24bf-f881-3f96-8789-551a23df7ad5,bachrach2007computing,10.1007/s10458-008-9057-6,resnick2009cost}.

\subsection{Cooperative Games and the Banzhaf Value}
We begin by formally defining a cooperative game with transferable utility (TU-game)
\begin{definition}[TU-game]
A  (TU game) is described by  a pair $(N, v)$ where $N = \{1, 2, \ldots, n\}$ is a finite set of $n$ players, and $v: 2^N \rightarrow \mathbb{R}$ is the characteristic function. This function assigns a real-valued worth $v(C)$ to each coalition $C \subseteq N$, with the convention that $v(\emptyset) = 0$.
\end{definition}

The characteristic function $v$ captures the value coalition $C$ generates collectively. A central objective in cooperative game theory is to formulate a solution concept—a principle for distributing the total value generated, $v(N)$, among the players. The Banzhaf value is a prominent solution concept that quantifies player influence based on their average contribution. 

\begin{definition}[Banzhaf Value] \footnote{It is important to distinguish the Banzhaf value from the Banzhaf Power Index. The Banzhaf Power Index normalises the Banzhaf value such that the sum of all players' indices equals 1, and is specifically used for simple games with binary outcomes (0/1). In contrast, the Banzhaf value we employ measures absolute marginal contributions in games with real-valued outcomes, making it appropriate for quantifying the impact of agents on network flow magnitude.}

For a TU-game $(N, v)$, the marginal contribution of a player $j \in N$ to a colaition $C \subseteq N \setminus {j}$ is defined as:
$\Delta_j(C) = v(C \cup \{j\}) - v(C)$.

The Banzhaf value $\beta_j(v)$ for player $j$ is the expectation of their marginal contribution, taken over all coalitions they are not a part of:
\begin{equation}
\beta_j(v) = \frac{1}{2^{n-1}} \sum_{C \subseteq N \setminus {j}} [v(C \cup {j}) - v(C)]
\end{equation}
\end{definition}
\subsection{Cardinal Network Flow Games}

\begin{definition}[Network Flow Game Setup]
Let $G = (V, E)$ be a directed graph with a source node $s \in V$ and a sink node $t \in V$. Let $c: E \rightarrow \mathbb{R}$ be a capacity function assigning a non-negative capacity to each edge. The set of players is a finite set of agents $A = \{a_1, a_2, \ldots, a_m\}$. The control of edges is defined by a surjective mapping $\text{ctrl}: E \rightarrow A$, where $\text{ctrl(e)}$ denotes the agent controlling edge $e$.
\end{definition}

From this setup, we define the Cardinal Network Flow Game.

\begin{definition}[Cardinal Network Flow Game]

A cardinal network flow game is the TU game $(A, v_G)$ where:
\begin{enumerate}
    \item The set of players is the set of agents $A = \{a_1, a_2, \ldots, a_m\}$
    \item For any coalition $C \subseteq A$, the characteristic function $v_G(C)$ is the value of the maximum flow from $s$ to $t$ in the subgraph $G_C = (V, E_C)$, where the edge set is restricted to those edges controlled by agents in the coalition: $E_C = \{e \in E | \text{ctrl}(e)\in C\}$. The value is thus:
    \begin{equation}
    v_G(C) = \text{maxflow}(G_C) 
    \end{equation}
\end{enumerate}
\end{definition}

This definition must be distinguished from the alternative formulation known as the Threshold Network Flow Game. In such a game, a flow quota  $q \geq 0$, is introduced, and the characteristic function $v_q(C)$ is defines as:
\begin{equation}
v_q(C) = \begin{cases}
1, & \text{if } \text{maxflow}(G_C) \geq q \\
0, & \text{otherwise}
\end{cases}
\end{equation}
The threshold variant constitutes a simple game, for which the Banzhaf Power Index is the appropriate measure. Our focus on the quantitative contribution of agents to the total network throughput necessitates the analysis of the cardinal game, thus requiring the use of the Banzhaf value.

For readers unfamiliar with network flow games, we provide a detailed worked example in Appendix~\ref{appendix:example}, demonstrating how coalition values are computed and how agent influence emerges from network topology.

\section{Methodology}
Our research employs a comprehensive approach to approximating power indices using GNN, focusing on generating diverse and representative datasets that capture the complexity of real-world systems.

\subsection{Problem formulation}
We frame the task as learning a function $f:G\rightarrow \mathbb{R}^m$, where $G$ is a network flow game instance and $m$ is the number of agents. The function $f$ takes the graph structure and agent configuration as input and outputs a vector containing the predicted Banzhaf value for each of the $m$ agents as output. 
To make this compatible with GNNs, we represent each game instance using a standard graph data structure with node and edge features:

\begin{enumerate}
    \item Node features: Each node is assigned a categorical type based on its structural role: source nodes (type 0), intermediate nodes (type 1), and sink nodes (type 2). These integer indices are embedded into dense vectors during training.
    \item Edge features: Each edge $e$ is represented by a feature vector of dimension $m+1$: 
    \begin{itemize}
        \item Edge capacity: $c(e) \sim \mathcal{U}(1,2,\dots,10)$ and normalised to z-scores within each graph instance, resulting in values in the range [-2,2].
        \item Agent control: A one-hot vector of length $m$ indicating which agent controls the edge
    \end{itemize}
    For example, an edge with normalised capacity $z=1.29$ controlled by an agent in a 5-agent game would be encoded as [1.29,0,1,0,0,0]
    \item Degree features: For each node, we compute in-degree and out-degree as auxiliary structural features.
    \item Output(Label): The target for each graph is an m-dimensional vector $\beta \in \mathbb{R}^m$, where $\beta_j$ is the normalized Banzhaf value of the agent $a_j$, with $\sum_{j=1}^{m}{\beta_j=1}$.
\end{enumerate}

This formulation allows the GNN to learn a mapping from the explicit graph and agent structure to the implicit, game-theoretic influence of each agent.

\subsection{Dataset generation strategies}
\label{data_generation_strategies}

We generated a large synthetic dataset of directed network flow games using the Erdős-Renyi algorithm~\cite{erdos59a}. Our datasets were created with numerous different structural properties of the games, to allow us to test the generalisation across different configurations. The generation processes, detailed in Algorithm \ref{alg:data_generation}, are governed by three main parameters: the number of nodes (n), the number of agents (m) and the edge probability (p).

For our experiments, we generated datasets for a grid of parameter values:
\begin{enumerate}
    \item Number of nodes ($n$) $\in \{20,50,100\}$
    \item Number of agents ($m$): $\in \{5,10,20\}$\footnote{The main power index calculation bottleneck doesn't lie in the graph size, which has a complexity of $O(E*V2)$~\cite{10-1145-321694-321699}, but rather in the number of agents, which has a complexity of $2^m$~\cite{415dc80a-289c-39e1-8e6b-601fc5ef267e}}.
    \item Edge probability ($p$): $\in \{0.5, 0.6, 0.7, 0.8, 0.9, 1.0\}$
\end{enumerate} 

Each configuration comprises 200,000 instances(80:20 train/test split), providing sufficient data for deep learning models to learn the distribution effectively, as Neural Networks require large amounts of data to perform optimally~\cite{Bishop1995}.

\begin{algorithm}[h]
\refstepcounter{algorithm}
\addtocounter{algorithm}{-1}  
\caption{Dataset generation}
\label{alg:data_generation}
\begin{algorithmic}[1]
\Require Dataset size $k$, max edge capacity $c$, Number of vertices $n$, list of agents $A=[a_1, a_2, \dots\ a_m]$, Edge Probability $p_{edge}$, source node $s$, sink node $t$ 
\Ensure Network flow graph $G$
\For{$i = 1$ to $k$}
    \State $G \gets \textproc{Erdős-Renyi}(n, p_{edge})$
    \State $G \gets \textproc{RemoveIncomingEdges}(G, s)$ \Comment{Ensure that the source is strong}
    \State $G \gets \textproc{RemoveOutgoingEdges}(G, t)$ \Comment{Ensure that the sink is strong}
    \ForAll{$e \in E(G)$}
        \State $capacity(e) \sim \mathcal{U}(1, 2, \dots, c)$
        \State Assign $agent(e) \sim A$
    \EndFor
\EndFor
\State \Return $G, A$

\end{algorithmic}
\end{algorithm}

\subsection{Ground-truth label generation}
As the exact computation of the Banzhaf value is intractable for $m=20$ due to its complexity ($\mathcal{O}(2^m)$), we generated ground-truth labels using a hybrid approach: exact computation via complete coalition enumeration for $m\in \{5,10\}$, and high-fidelity Monte-Carlo approximation ($N=10,000$ samples) for $m =20$. This ensures maximum accuracy for smaller instances while maintaining computational feasibility for larger agent sets.

The Monte Carlo approximation procedure, detailed in Algorithm \ref{alg:banzhaf_calc}, estimates the Banzhaf value for every agent based on their average marginal contribution across randomly sampled coalitions.
For each of the $N=10,000$ randomly sampled subgraphs, we compare whether the maximal flow from the source to the sink is changed in the subgraph when the edges of an agent $a$ are removed from the original subgraph. The absolute difference for each agent is accumulated. Finally, the vector of total marginal contributions for all agents is normalised to the sum of 1, yielding each agent's relative share $\beta_a$ of the total influence in the game. This normalised vector serves as the ground-truth label for our supervised learning task.

\begin{algorithm}[h]
\caption{Monte-Carlo Banzhaf Value Approximation for Network Flow Games}
\label{alg:banzhaf_calc}
\begin{algorithmic}[1]
\Require Graph $G=(V,E,c,\text{ctrl})$, agents $A=\{a_1, \dots, a_m\}$, number of samples $N$
\Ensure Normalized Banzhaf value vector $\beta \in \mathbb{R}^m$

\State Initialize $M_j \gets 0$ for all $j \in \{1, \dots, m\}$ \Comment{Sum of marginal contributions}

\For{$i = 1$ to $N$}
    \State $C \gets \textproc{SampleRandomCoalition}(A)$ \Comment{e.g., a random binary vector}
    \State $v_C \gets \textproc{MaxFlow}(G, C)$ \Comment{Max flow using edges of agents in $C$}
    
    \For{agent $a_j \in A$}
        \State $C' \gets C \oplus \{a_j\}$ \Comment{Toggle agent $a_j$'s membership in $C$ (XOR)}
        \State $v_{C'} \gets \textproc{MaxFlow}(G, C')$
        \State $M_j \gets M_j + |v_C - v_{C'}|$ \Comment{Accumulate absolute marginal contribution}
    \EndFor
\EndFor

\State TotalContribution $\gets \sum_{j=1}^m M_j$
\If{TotalContribution $> \epsilon$} \Comment{Avoid division by zero}
    \State $\beta_j \gets M_j / \text{TotalContribution}$ for all $j \in \{1, \dots, m\}$
\Else
    \State $\beta_j \gets 1/m$ for all $j \in \{1, \dots, m\}$ \Comment{Assign uniform power if no flow}
\EndIf

\State \Return $\beta$
\end{algorithmic}
\end{algorithm}

\subsection{Data Availability}
To ensure reproducibility and support future research, we will publicly release the dataset upon paper acceptance. The initial release comprises 51 configurations (\~500GB, 10.2M graphs) spanning all experimental conditions except $n=100, m=20, p \in \{0.6, 0.8, 0.9\}$, which could not be generated within the manuscript preparation timeline. We intend to complete and add these configurations to the public dataset as computational resources permit.

\subsection{Model training}
We evaluated a suite of prominent GNN architectures to determine their suitability for the task:
\begin{enumerate}
    \item Graph Attention Networks (GAT) ~\cite{veličković2018graphattentionnetworks}.
    \item Graph Isomorphism Network with Edge features (GINE) \cite{xu2019powerfulgraphneuralnetworks}
    \item Graph Convolution Network with Edge-features(EdgeConv) \cite{gilmer2017neuralmessagepassingquantum}
    Note: We refer to this architecture as EdgeConv for brevity, though it uses NNConv (Neural Network Convolution) layers.
\end{enumerate} 
All models were trained with an AdamW optimiser for 100 epochs with early stopping based on validation performance to prevent overfitting. We employed the Huber loss as the objective function due to its robustness to small variations in the target values. For networks with many agents (e.g., m=20), individual Banzhaf values can be very small (averaging 0.05), making standard MSE or MAE less suitable for capturing meaningful relative differences between agents' influence.

Our evaluation protocol was designed to assess two distinct capabilities:
\begin{enumerate}
    \item In-distribution performance: For each dataset configuration (a specific combination of n,m,p values), we train the model on the 80\% training set and evaluate on the remaining 20\% test set to test its predictive accuracy.
    \item Zero-shot generalisation: To test the model's ability to generalise, we evaluated models trained on one configuration (e.g., n=50,m=10,p=0.5) on a different configuration's test set (e.g. n=100, m=20, p=0.7) without any fine-tuning. This rigorously tests the model's capability to learn underlying principles of agent influence beyond the specific distribution of its training data.
\end{enumerate}

To evaluate generalisation across different agent counts, we used zero-padded features to maintain architectural compatibility (see Appendix \ref{appendix:protocol_and_architectures} for exact evaluation and model implementation details).

\section{Results}
We first evaluate in-distribution performance (same generation parameters for train and test sets), then assess out-of-distribution generalisation (different generation parameters between train and test sets)

Note that absolute loss values naturally decrease for configurations with higher agent count ($m$), as the Banzhaf values are smaller on average. Our analysis focuses on relative performance and generalisation trends.

\subsection{In-distribution}
\begin{figure}[htbp]
    \centering
      \begin{subfigure}{0.95\textwidth}
        \centering
        \begin{subfigure}{0.3\textwidth}
            \centering
            \includegraphics[width=\textwidth]{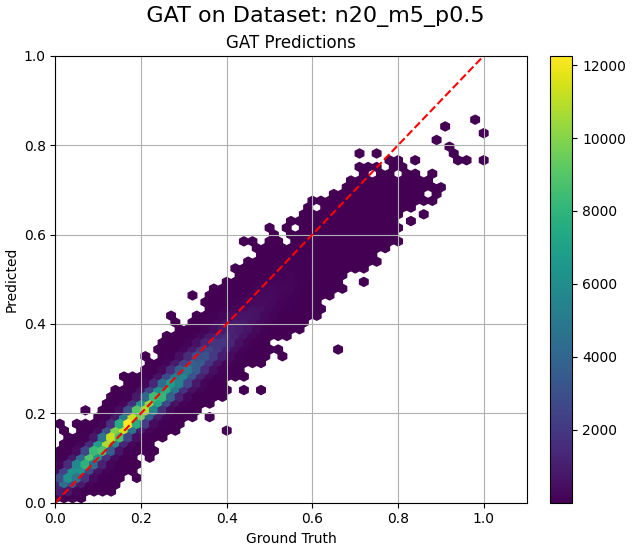}
            \label{fig:sub7}
        \end{subfigure}
        \hfill
        \begin{subfigure}{0.3\textwidth}
            \centering
            \includegraphics[width=\textwidth]{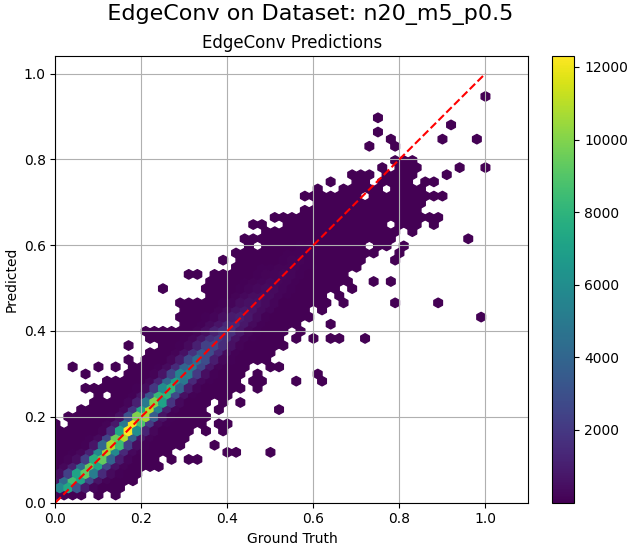}
            \label{fig:sub8}
        \end{subfigure}
        \hfill
        \begin{subfigure}{0.3\textwidth}
            \centering
            \includegraphics[width=\textwidth]{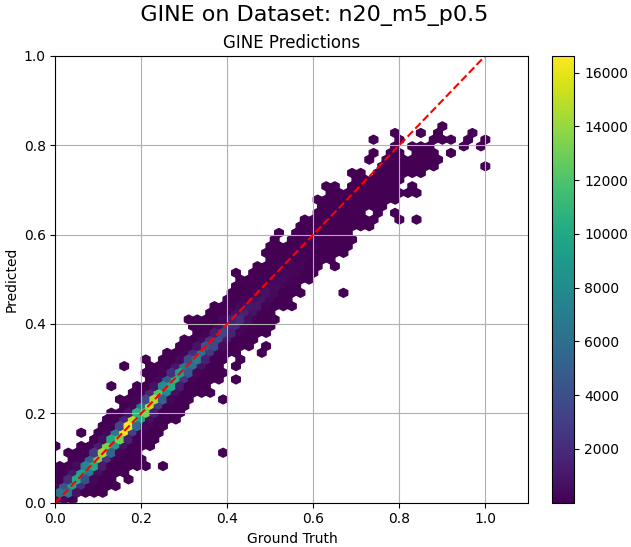}
            \label{fig:sub9}
        \end{subfigure}
        \caption{In-distribution prediction performance on the n=20, m=5, p=0.5 test set for the GAT (left), EdgeConv (centre), and GINE (right) models.}
        \label{fig:row3}
    \end{subfigure}
    
    \medskip
    
      \begin{subfigure}{0.95\textwidth}
        \centering
        \begin{subfigure}{0.3\textwidth}
            \centering
            \includegraphics[width=\textwidth]{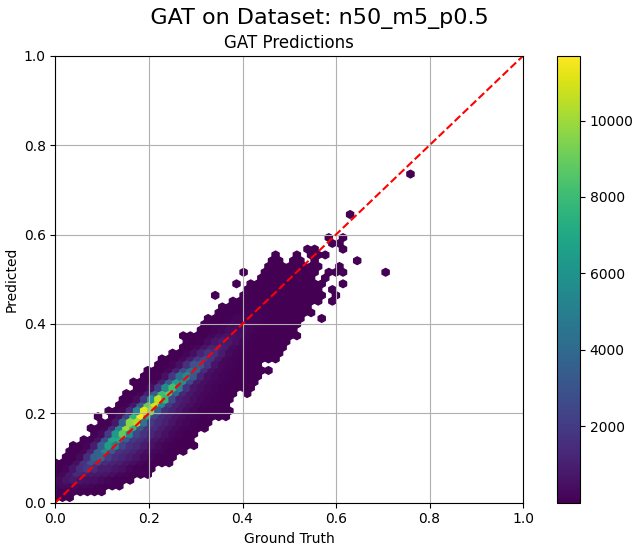}
            \label{fig:sub7}
        \end{subfigure}
        \hfill
        \begin{subfigure}{0.3\textwidth}
            \centering
            \includegraphics[width=\textwidth]{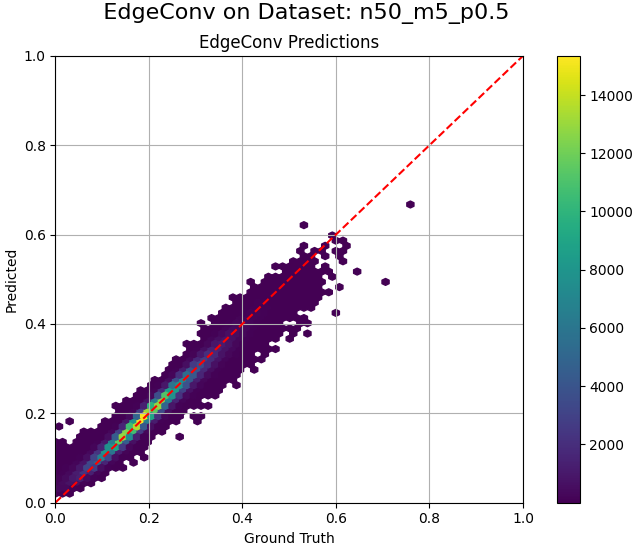}
            \label{fig:sub8}
        \end{subfigure}
        \hfill
        \begin{subfigure}{0.3\textwidth}
            \centering
            \includegraphics[width=\textwidth]{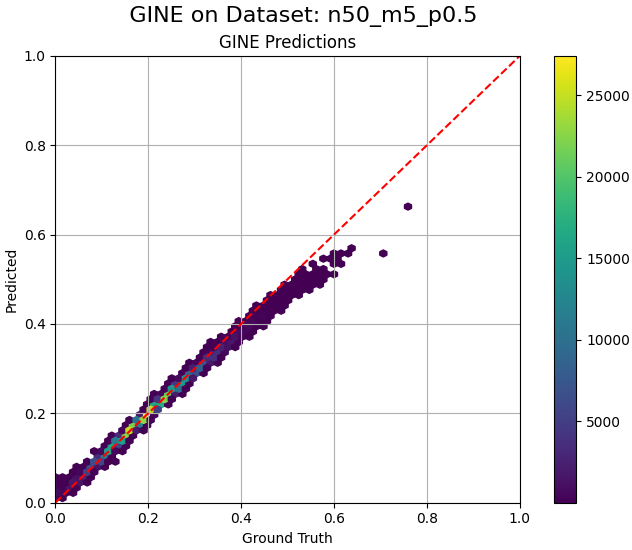}
            \label{fig:sub9}
        \end{subfigure}
        \caption{In-distribution prediction performance on the n=50, m=5, p=0.5 test set for the GAT (left), EdgeConv (centre), and GINE (right) models.}
        \label{fig:row3}
    \end{subfigure}
    
    \medskip
    
      \begin{subfigure}{0.95\textwidth}
        \centering
        \begin{subfigure}{0.3\textwidth}
            \centering
            \includegraphics[width=\textwidth]{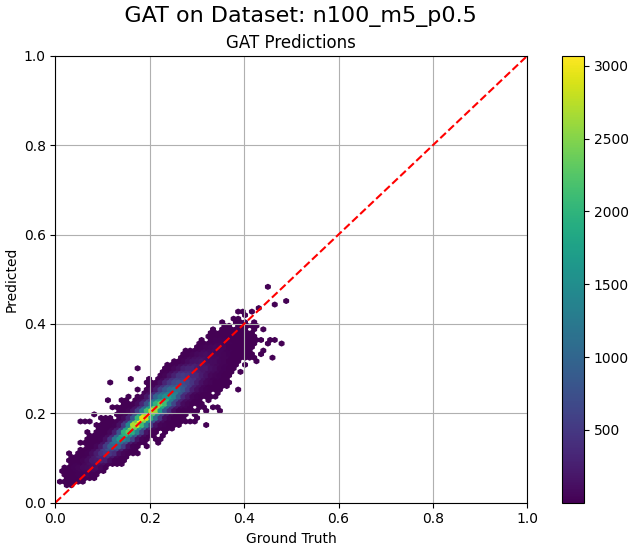}
            \label{fig:sub7}
        \end{subfigure}
        \hfill
        \begin{subfigure}{0.3\textwidth}
            \centering
            \includegraphics[width=\textwidth]{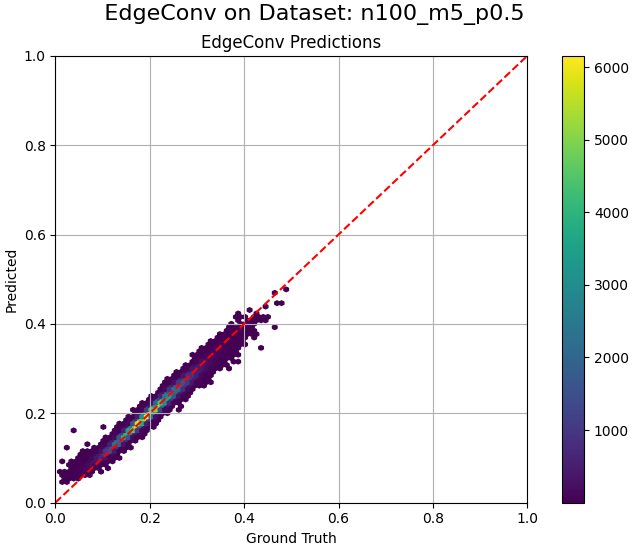}
            \label{fig:sub8}
        \end{subfigure}
        \hfill
        \begin{subfigure}{0.3\textwidth}
            \centering
            \includegraphics[width=\textwidth]{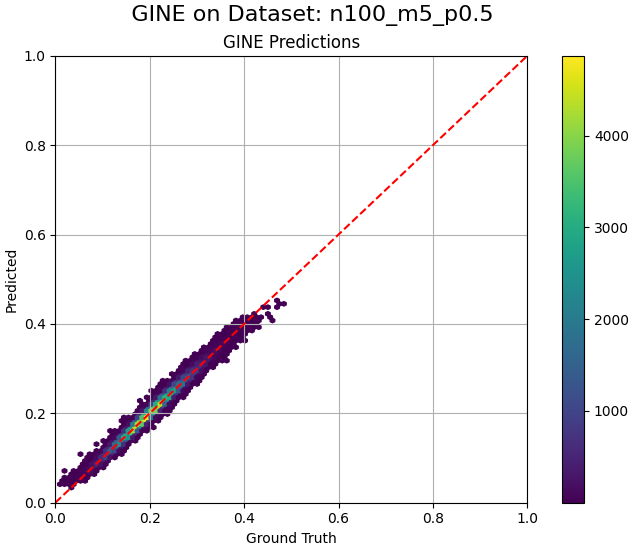}
            \label{fig:sub9}
        \end{subfigure}
        \caption{In-distribution prediction performance on the n=100, m=5, p=0.5 test set for the GAT (left), EdgeConv (centre), and GINE (right) models.}
        \label{fig:row3}
    \end{subfigure}
    
    \caption{Hexplot comparison of three model architectures on 5 agents on different-shaped graphs. Figures show graphs of sizes n=20, n=50 and n=100 for rows a,b and c, respectively. All of the edge probabilities are equal to p=0.5, with the horizontal axis representing the ground truth Banzhaf values, and the vertical axis the model's prediction. The colour indicates the point density (from low density(purple) to high density (yellow). The red dashed line indicates a perfect prediction.}
    \label{fig:architecture_in_distribution_model_comparison}
\end{figure}

Figure \ref{fig:architecture_in_distribution_model_comparison} presents a qualitative comparison of the GAT, EdgeConv and GINE models, trained and evaluated on a representative dataset of graphs with 5 agents, an edge probability of 0.5 and 20,50, and 100 nodes (represented in the first, second and third row, respectively). The plots correlate the ground-truth Banzhaf values (horizontal axis) with the model's prediction (Vertical axis). A perfect prediction would place all points directly on the diagonal red line.

Visually, all architectures demonstrate a strong ability to learn the Banzhaf value, capturing the linear relationship with high fidelity, as indicated by the dense clustering of points along the diagonal. However, a closer inspection reveals that the GINE architecture exhibits the tightest distribution around the line, with fewer extreme outliers, suggesting a more robust and accurate mapping. The colour density indicates that all models are particularly accurate for the more common, smaller Banzhaf values (Which, for graphs with 5 agents, would be around a Banzhaf value of 0.2 per agent). 

\subsection{Cross edge probability evaluations}
\begin{figure}[h]
     \centering
     \includegraphics[width=0.95\textwidth]{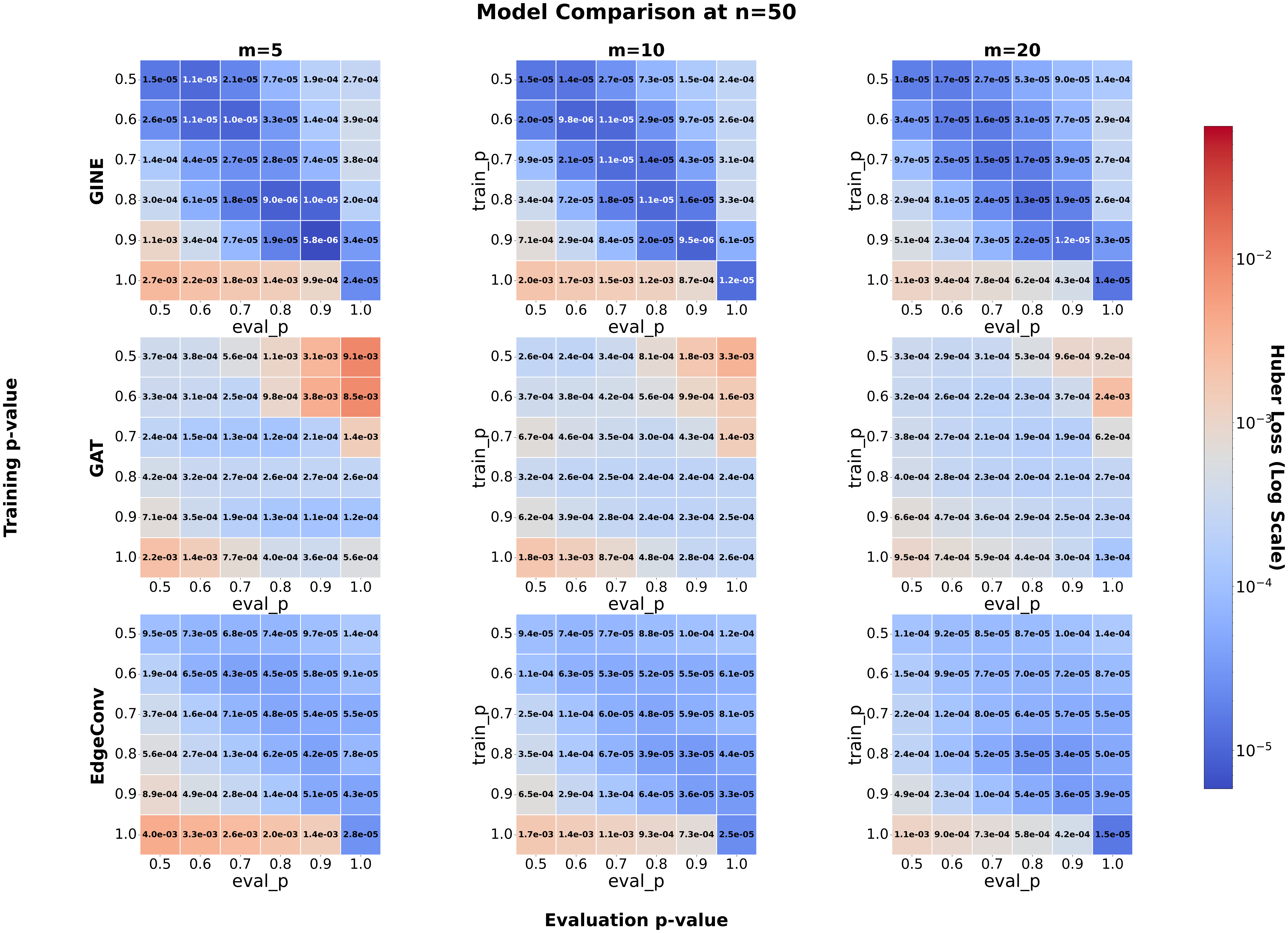}
     \caption{Cross-architecture edge-probability generalisation performance at $n=50$. Heatmaps show Huber loss for models trained and evaluated at different edge probabilities (p). Columns represent different agent counts ($m=5, 10, 20$). Each cell displays performance when a model trained at a given p-value (y-axis) is evaluated at another p-value (x-axis). Models trained at p=1.0 exhibit poor generalisation across all architectures, while optimal training regimes differ by architecture: $p=0.5$ for GINE, $p=0.8$ for GAT, and $p=0.6$ for EdgeConv.}
     \label{fig:cross_p_all}
\end{figure}

When looking at the out-of-distribution performance, architectural differences become more pronounced (Figure \ref{fig:cross_p_all}). Both GINE and EdgeConv demonstrate robust generalisation across lower edge probabilities, whereas GAT exhibits consistently higher loss values and shows asymmetric generalisation: models trained on lower p-values fail to generalise to higher p-values, unlike the other architectures. Notably, models trained at $p=1.0$ fail to generalise to other configurations. This is attributable to the structural simplicity of fully connected graphs: when $p=1.0$, all intermediate nodes connect directly to both source and sink, reducing maximal flow calculations to trivial two-hop paths. This simplified topology inadequately represents the complexity of agent contributions in sparser graphs, leaving models unprepared for more realistic network structures. Importantly, this asymmetry is architecture-dependent: GINE and EdgeConv models trained on sparse graphs $p<1.0$ successfully generalise to the dense $p=1.0$ case, while GAT does not.

These results suggest that training on intermediate sparsity regimes ($p \approx 0.5-0.7$) provides the most robust generalisation across graph topologies, with GINE emerging as the most reliable architecture for varying network densities. Given GINE's superior performance and generalisation capabilities, we conduct a detailed analysis of this architecture across all experimental configurations.

\begin{figure}[h]
     \centering
     \includegraphics[width=0.95\textwidth]{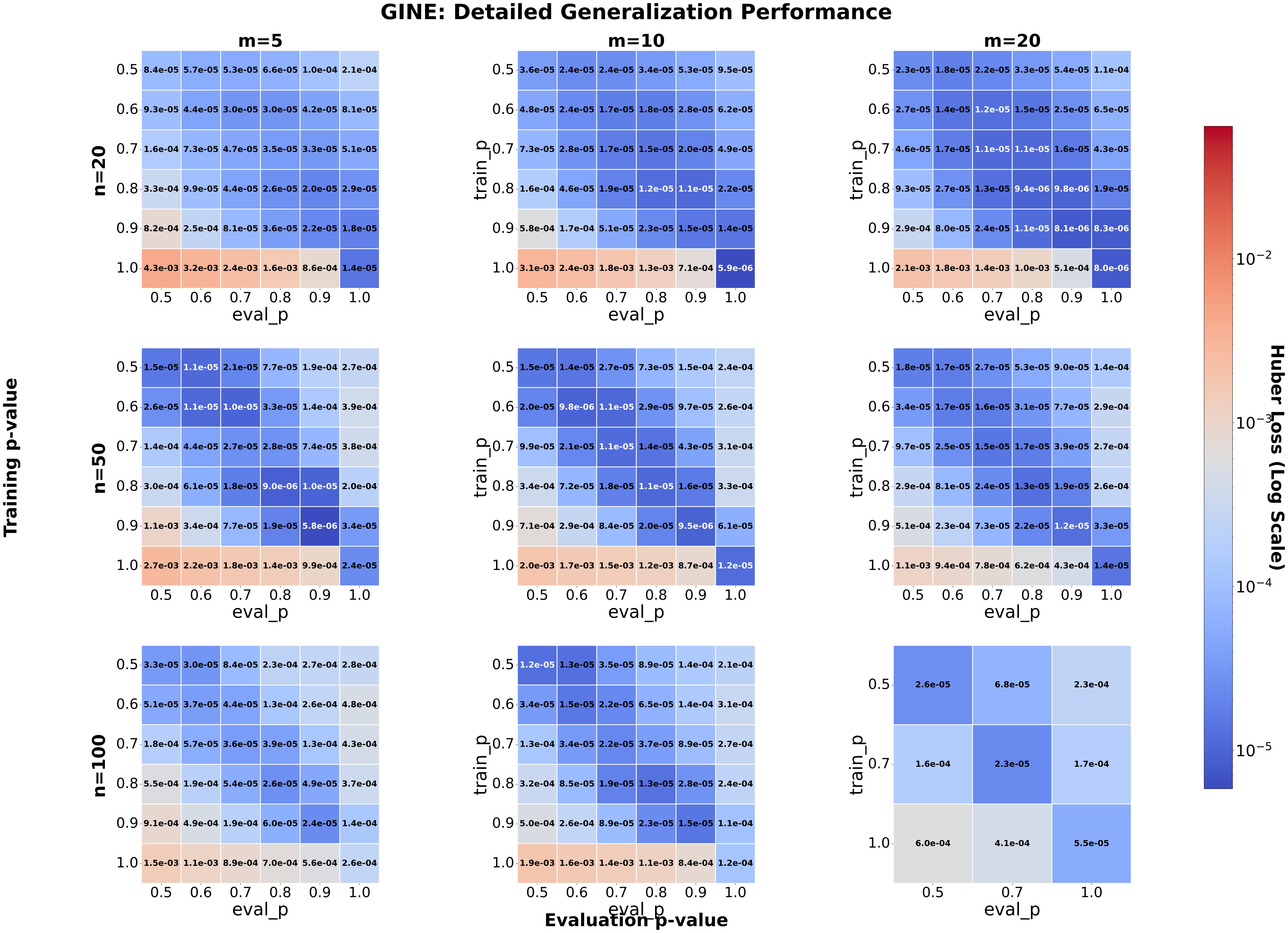}
     \caption{Detailed edge-probability generalisation analysis for GINE across all training configurations. Heatmaps show Huber loss (log scale) as a function of training edge probability (train\_p, y-axis) and evaluation edge probability (eval\_p, x-axis) for different graph sizes ($n=20, 50, 100$; rows) and agent counts ($m=5, 10, 20$; columns). Diagonal elements consistently achieve the lowest error (in-distribution), while off-diagonal performance degrades more severely for larger n values, indicating reduced generalisation capacity in larger graphs.}
     \label{fig:cross_p_gine}
\end{figure}

Focusing on GINE (Figure \ref{fig:cross_p_gine}), the model generalises across all graph sizes ($n$), though performance degrades for larger networks. Note that results for $n=100, m=20$ at $p\in \{0.6,0.8,0.9\}$ are unavailable due to dataset generation constraints. Interestingly, loss values appear lower for higher agent counts ($m$), but this is an artefact of normalisation: Banzhaf values are normalised by the number of agents, naturally yielding smaller absolute values for larger $m$. The raw generalisation capacity remains comparable across agent counts. The heatmaps reveal that training at intermediate sparsity values ($p \approx 0.6-0.7$) provides the most balanced generalisation across all edge probabilities, with particularly strong performance maintained even when generalising to $p=1.0$.

Detailed cross-p analysis for GAT and EdgeConv are provided in Appendix \ref{appendix:gat_full_results} and \ref{appendix:edgeconv_full_results}, respectively.

Based on these findings, and considering the computational cost of generating the full density spectrum for the largest configurations, we focus subsequent cross-graph size and cross-agent count analyses on three representative edge probabilities: $p=0.5$ (sparse), $p=0.7$ (intermediate), and $p=1.0$ (fully connected). These values capture the full range of network densities while revealing distinct architectural behaviours, as demonstrated above.

\subsection{Cross graph size evaluations}
\begin{figure}[h]
     \centering
     \includegraphics[width=0.95\textwidth]{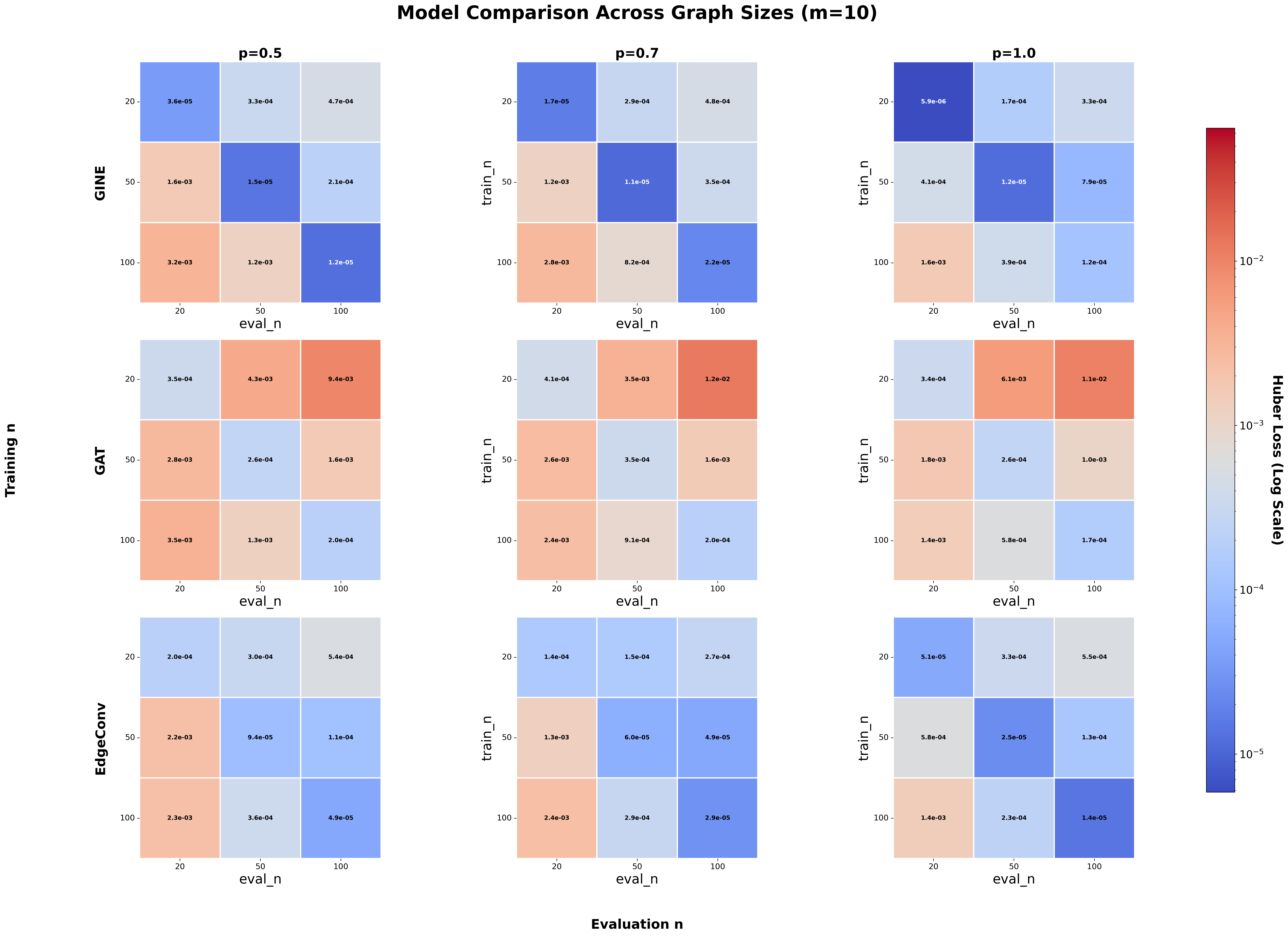}
     \caption{Cross-architecture graph size generalisation performance at $m=10$. Heatmaps show Huber loss for models trained and evaluated at different numbers of nodes (n). Columns represent different edge probabilities ($p=0.5, 0.7, 1.0$). Each cell displays performance when a model trained at a given n-value (y-axis) is evaluated at another n-value (x-axis). Models trained at n=100 exhibit poor generalisation across all architectures, while optimal training regimes differ by architecture: $n=20$ for GINE and EdgeConv, $n=50$ for GAT.}
     \label{fig:cross_n_all}
\end{figure}

Architectural differences persist for out-of-distribution performance on different graph sizes (Figure \ref{fig:cross_n_all}). All models maintained the best performance when evaluated on graphs of the same distribution as those they were trained on. However, all models additionally show a directional generalisation capability, namely, models that were trained on larger graphs perform worse when evaluated on smaller ones, whereas models trained on smaller graphs performed significantly better on larger graphs than the reverse. GAT seems to be unique in showing poor generalisation in both directions, particularly when training at extreme values ($n=20$ or $n=100$)

The results suggest that training on larger graphs is a suboptimal solution for model generalisation, and that if generalisation is the goal, training on smaller graphs allows the model to understand and extrapolate better for larger graphs, with GINE emerging once more as the most reliable architecture for varying network sizes. However, generalisation capacity diminishes with increasing size disparity: models trained on $n=20$ perform worse on $n=100$ graphs compared to models trained on $n=50$.

\begin{figure}[h]
     \centering
     \includegraphics[width=0.95\textwidth]{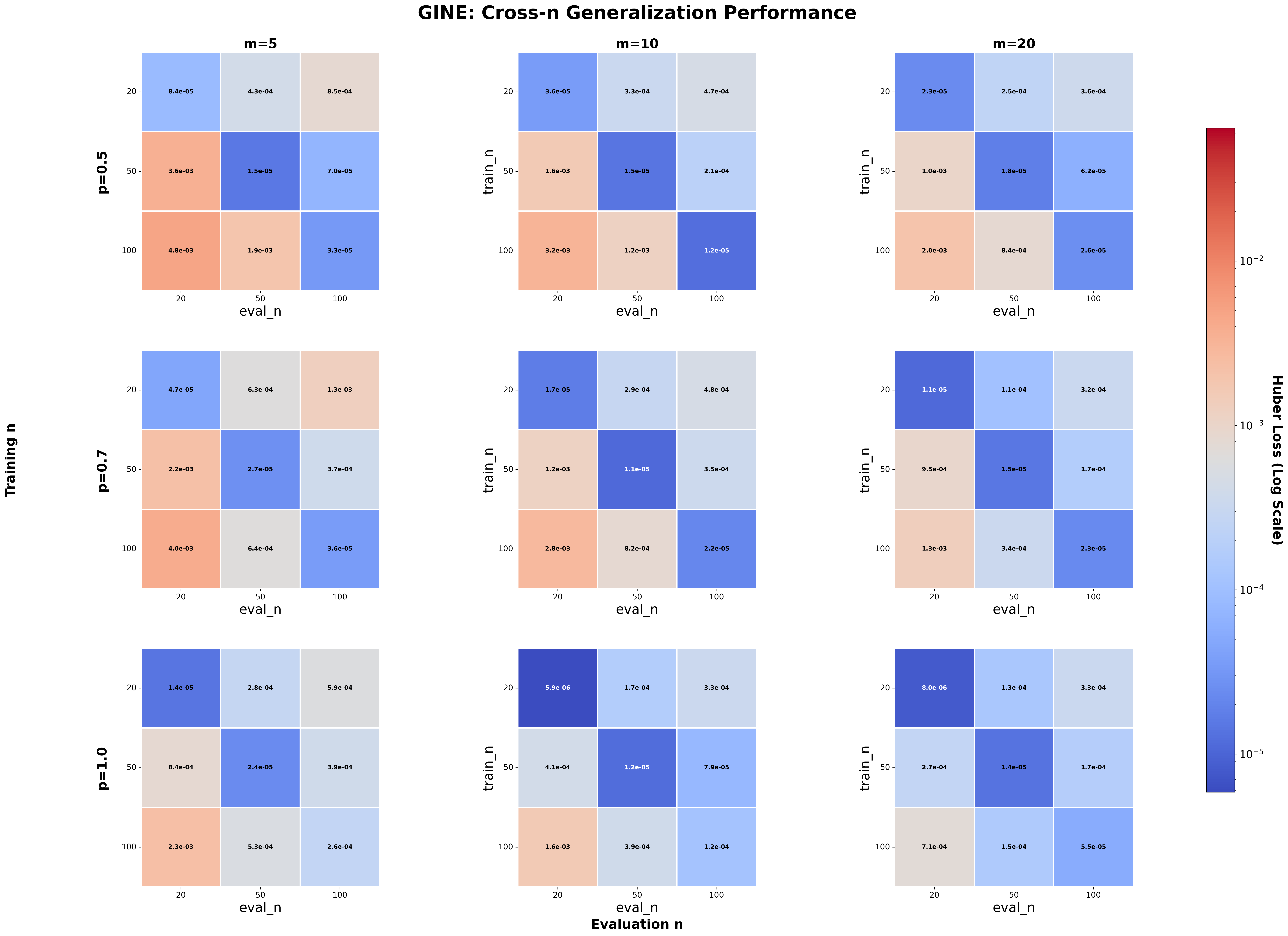}
     \caption{Detailed graph size generalisation analysis for GINE across all training configurations. Heatmaps show Huber loss (log scale) as a function of training number of nodes (train\_n, y-axis) and evaluation number of nodes (eval\_n, x-axis) for different edge probabilities ($p=0.5, 0.7, 1.0$; rows) and agent counts ($m=5, 10, 20$; columns). Diagonal elements consistently achieve the lowest error (in-distribution), while off-diagonal performance degrades more severely for larger n values, indicating reduced generalisation capacity in larger graphs.}
     \label{fig:cross_n_gine}
\end{figure}

Closer examination of GINE (Figure \ref{fig:cross_n_gine}) reveals that cross-size generalisation is strongest for dense graphs ($p=1.0$). This capability to generalise across graph sizes but not densities is surprising and suggests two possible explanations: either max flow estimation exhibits greater structural consistency across scales in dense networks compared to sparse ones, or the uniform structure of $p=1.0$ graphs forces the model to learn more abstract, generalisable features that transfer better across sizes. The persistent inability of $p=1.0$-trained models to generalise downward to smaller graphs remains unexplained and warrants further investigation. 

Detailed cross-n analysis for GAT and EdgeConv are provided in Appendix \ref{appendix:gat_full_results} and \ref{appendix:edgeconv_full_results}, respectively.

\subsection{Cross agent count  evaluations}
\begin{figure}[h]
     \centering
     \includegraphics[width=0.95\textwidth]{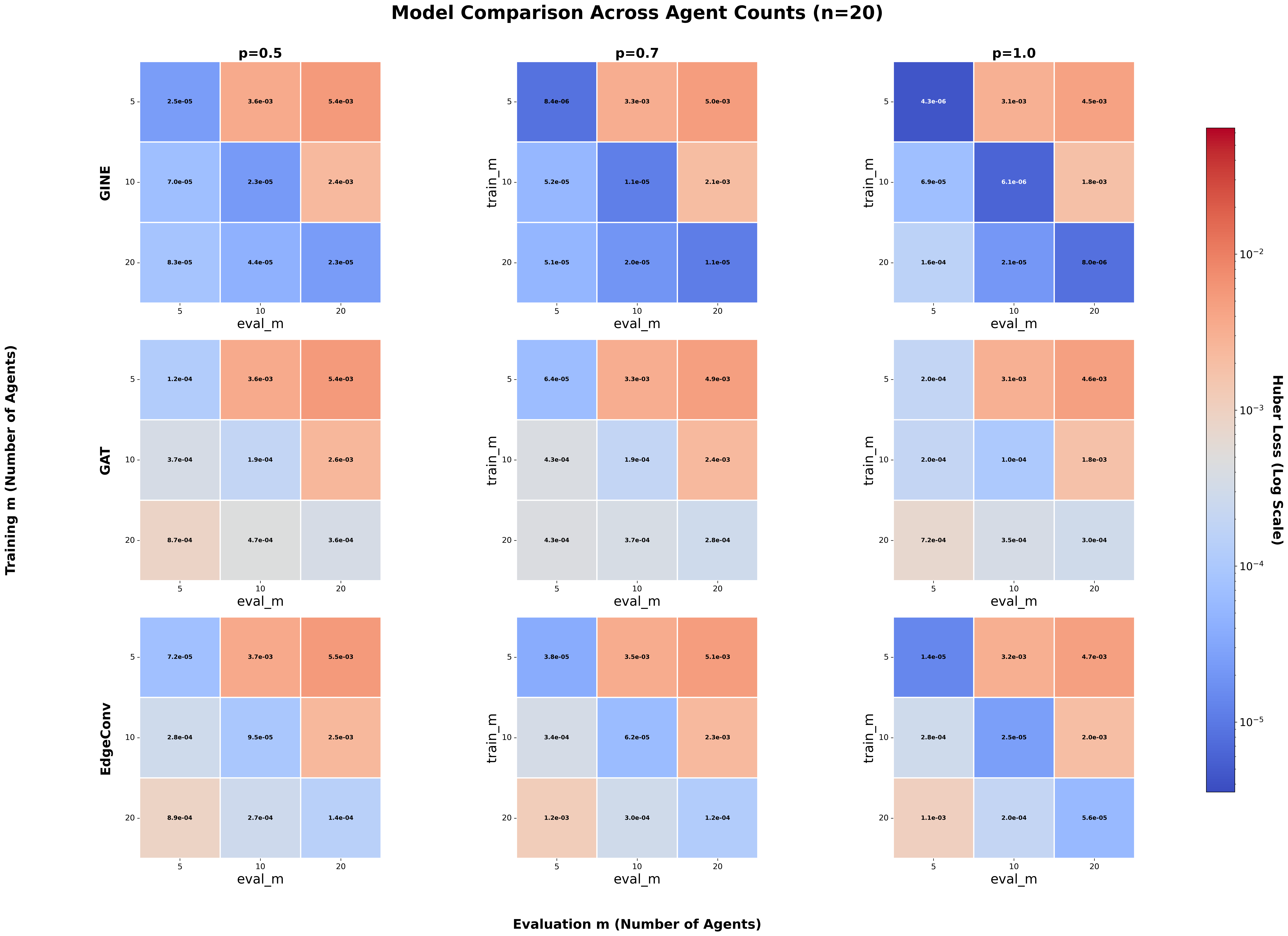}
     \caption{Cross-architecture agent count generalisation performance at $n=20$. Heatmaps show Huber loss for models trained and evaluated at different number of agents (m). Columns represent different edge probabilities ($p=0.5, 0.7, 1.0$). Each cell displays performance when a model trained at a given m-value (y-axis) is evaluated at another m-value (x-axis). Models trained at m=5 exhibit poor generalisation across all architectures, while optimal training regimes are mutual to architecture at $m=20$.}
     \label{fig:cross_m_all}
\end{figure}

Architectural differences persist for out-of-distribution performance across different agent counts (Figure \ref{fig:cross_m_all}). All models achieved best performance when evaluated on graphs of the same agent count as their training data. However, all models exhibit directional generalisation: models trained on higher agent counts generalise well to lower counts, whereas models trained on lower agent counts performed significantly worse when evaluated on higher agent counts.

Interestingly, this directional pattern is opposite to the cross-size generalisation observed earlier. While models trained on smaller graphs successfully generalise upward to larger graphs, models trained on fewer agents fail to generalise upward to more agents. This asymmetry suggests fundamentally different learning dynamics: graph size appears to teach scalable structural patterns (simpler that can be transferred to more complex cases), while agent count requires learning from richer strategic interactions (more complex can be transferred to the simpler case). Training with more agents may expose models to a fuller range of coalition dynamics and marginal contribution patterns, enabling downward transfer to simpler cases where fewer agents interact. It is important to note that this may be influenced by our zero-padding approach for varying agent counts; alternative masking methods might yield different generalisation characteristics. GINE again emerges as the most reliable architecture for cross-agent count generalisation. 

\begin{figure}[h]
     \centering
     \includegraphics[width=0.95\textwidth]{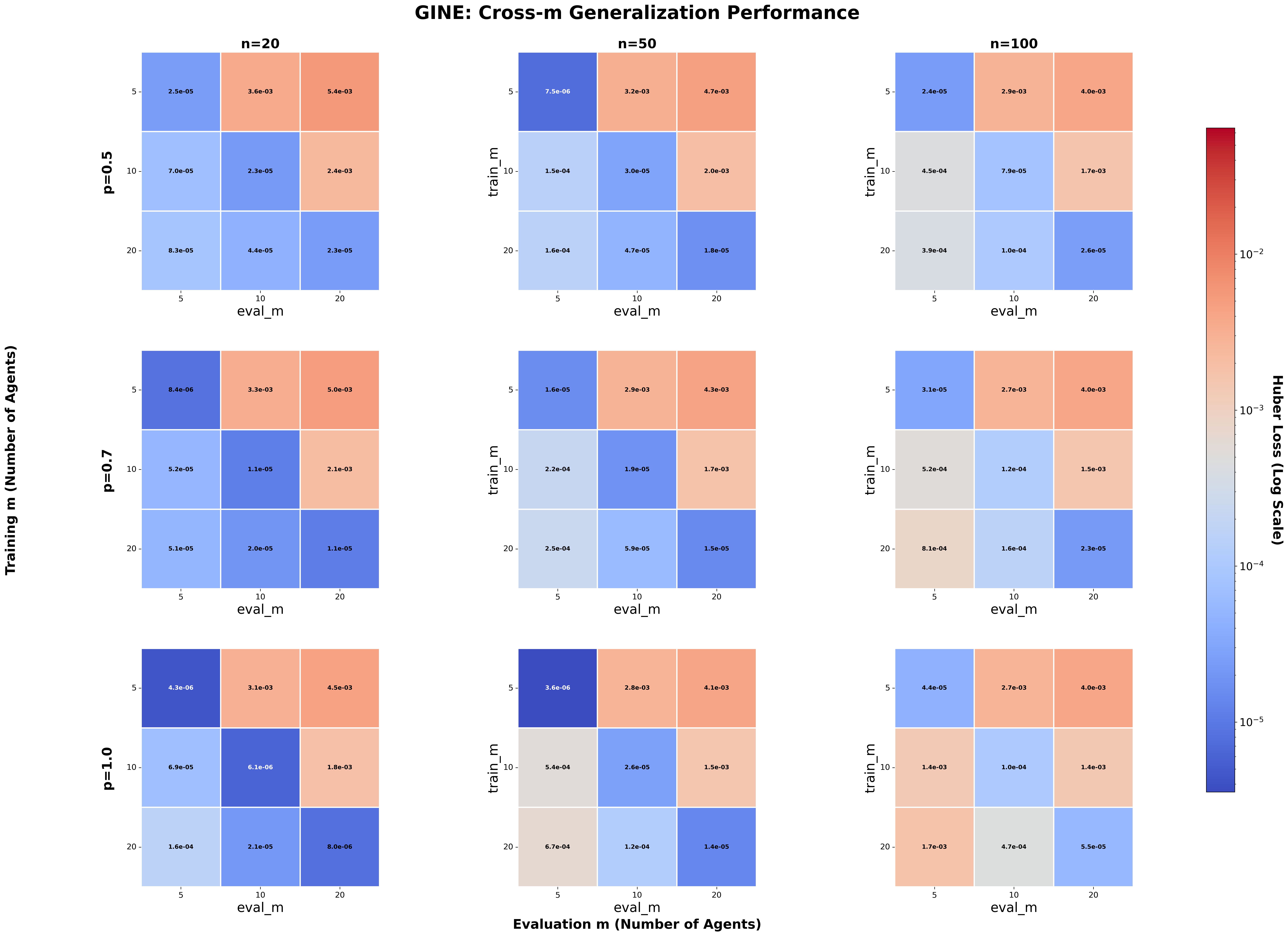}
     \caption{Detailed agent count generalisation analysis for GINE across all training configurations. Heatmaps show Huber loss (log scale) as a function of training number of nodes (train\_n, y-axis) and evaluation number of nodes (eval\_n, x-axis) for different edge probabilities ($p=0.5, 0.7, 1.0$; rows) and graph size ($n=20, 50, 100$; columns). Diagonal elements consistently achieve the lowest error (in-distribution), while off-diagonal performance degrades more severely for lower $m$ values, indicating reduced generalisation capacity in graphs.}
     \label{fig:cross_m_gine}
\end{figure}

Closer examination of GINE (Figure \ref{fig:cross_m_gine}) reveals that cross-agent count generalisation is strongest for smaller graphs ($n=20$) and denser topologies. The downward transfer pattern is particularly pronounced: models trained on $m=20$ maintain low error when evaluated on $m=5$ or $m=10$ across most configurations, while upward transfer consistently fails. This suggests that smaller, denser graphs provide a more stable learning environment where agent influence patterns are more clearly distinguished, enabling the model to learn transferable representations of coalition dynamics.

\section{Discussion}
Amongst tested architectures, GINE proved most robust, generalising effectively across graph sizes, edge densities, and agent counts. GAT surprisingly underperformed, failing to generalise across graph sizes and edge densities. The reasons for attention mechanisms' limitations here warrant further investigations. EdgeConv achieved good results but requires significantly longer training ($x10-20$ slower) for larger graphs due to its parameter scaling with graph size, though it remains preferable to GAT for smaller networks.

Key structural findings emerged across all architectures. Fully connected graphs ($p=1.0$) present a unique challenge: all nodes become direct source-sink connectors, making agent influence less pronounced than in sparser graphs requiring multi-hop paths. This structural simplification explains why $p=1.0$-trained models fail to generalise to other densities, yet paradoxically achieve the best cross-n and cross-m generalisation. We hypothesise that this uniform structure forces models to learn scale-invariant features transferable across sizes, even though these features miss the sparsity-dependent complexity needed for varying edge densities. 

The directional generalisation patterns reveal different learning dynamics for graph structure versus strategic complexity. Models trained on smaller graphs ($n=20$) generalise upward to larger graphs, learning fundamental flow patterns that scale effectively. Conversely, models trained on fewer agents ($m=5$) fail to generalise upward, suggesting that richer coalition dynamics in higher agent settings expose patterns absent in simpler cases. This asymmetry indicates that structural complexity(graph size) provides better inductive bias than strategic complexity (agent count) for generalisation. 

Practical deployment considerations favour GINE decisively. Beyond generalisation performance, GINE inference requires 0.6-0.7 milliseconds 
per graph across all configurations, compared to 0.19-20 seconds for Monte Carlo 
sampling ($N=10,000$)—yielding speedups from $x300$ (small networks) to $x32,000$ 
(large and dense networks). While training requires 2-12 hours for 200K graphs, successful zero-shot generalisation (demonstrated in the numerous cross-configuration evaluations) eliminates retraining costs for new topologies. Complete timing comparisons are provided in Appendix \ref{appendix:comp_analysis}.

Our architectural comparison advances understanding of both generalisation and power index approximation methods. Previous model-based approaches like FastSHAP and KernelBanzhaf focus primarily on sampling strategies. Our results demonstrate that architecture selection and training data scale enable robust cross-structure generalisation, suggesting these factors merit equal attention to sampling optimisation in power index approximation.

Several factors likely contribute to GINE's advantage in network flow games: GIN-based architectures' strong structural expressiveness aligns with tasks requiring global graph understanding; the observed training stability makes GINE more practical despite GAT's theoretical promise; and GINE's aggregation mechanism may interact more favourably with our feature encoding. Systematic ablation studies could disentangle these factors.

Our work establishes GNN-based approximation as a viable alternative to traditional sampling methods, particularly when repeated evaluations are needed. The amortised cost model, expensive training but near-instantaneous inference, shifts the computational trade-off space. We find that these models' value increases as the number of queries and configurations increases. 

Several limitations warrant acknowledgment. Our evaluation uses synthetic Erdős-Rényi graphs, which may not capture structural properties of real networks with community structure or hierarchical organization. Ground-truth labels for $m=20$ rely on Monte Carlo approximation rather than exact computation, potentially introducing label noise. Additionally, our zero-padding approach for varying agent counts may influence cross-m generalisation patterns. Future work should validate these findings on empirical networks, and investigate whether architectural insights transfer to other cooperative game solution concepts.

\bibliographystyle{unsrt}
\bibliography{refs} 

\newpage
\appendix

\section{Illustrative example of a Cardinal Network Flow Game}
\label{appendix:example}
To provide intuition for readers less familiar with network flow games, we present a detailed worked example illustrating how the characteristic function is computed and how agent influence emerges from network structure.

\subsection{Network setup}
Consider a simple network with four nodes: a source node $s$, two intermediate nodes $a$ and $b$, and a sink node $t$. The network is controlled by three agents $A = \{a_1, a_2, a_3\}$, where each edge is owned by exactly one agent who can choose whether to make it available for flow.

The complete network specification is given in Table~\ref{table:network_flow_example}. Agent $a_1$ controls the edges from the source to node $a$ (capacity 3) and the connection between the two intermediate nodes (capacity 1). Agent $a_2$ controls the edge from the source to node $b$ (capacity 2). Agent $a_3$ controls both edges leading to the sink: from node $a$ (capacity 2) and from node $b$ (capacity 3).

\begin{table}[h]
\caption{Network specification for illustrative example. Each edge has a maximum capacity and is controlled by one agent.}
\centering
\begin{tabular}{|c|c|c|}

\hline
Edge & Capacity & Controller \\ 
\hline
$(s,a)$ & 3 & $a_1$ \\
$(s,b)$ & 2 & $a_2$ \\
$(a,b)$ & 1 & $a_1$ \\
$(a,t)$ & 2 & $a_3$ \\
$(b,t)$ & 3 & $a_3$ \\
\hline
\end{tabular}
\label{table:network_flow_example}
\end{table}

\subsection{Computing Coalition Values}
The characteristic function $v_G(C)$ for any coalition $C$ is defined as the maximum flow from $s$ to $t$ using only edges controlled by agents in $C$. We now compute several coalition values to illustrate this concept.

\textbf{Coalition $C_1 = \{a_1, a_3\}$:} When agents $a_1$ and $a_3$ cooperate, the available edges are $E_{C_1} = \{e_1, e_3, e_4, e_5\}$. Notice that $e_2$ (controlled by $a_2$) is unavailable. In the subgraph $G_{C_1}$, flow can travel along two paths:
\begin{itemize}
    \item Path 1: $s \rightarrow a \rightarrow t$ with bottleneck capacity $\min(3,2) = 2$
    \item Path 2: $s \rightarrow a \rightarrow b \rightarrow t$ with bottleneck capacity $\min(3,1,3) = 1$
\end{itemize}
However, these paths share edge $e_1$, so we cannot send flow along both simultaneously at full capacity. The maximum flow is 2 units along path 1. Thus $v_G(\{a_1, a_3\}) = 2$.

\textbf{Coalition $C_2 = \{a_2, a_3\}$:} When agents $a_2$ and $a_3$ cooperate, the available edges are $E_{C_2} = \{e_2, e_4, e_5\}$. Note that edges $e_1$ and $e_3$ (both controlled by $a_1$) are unavailable. The only viable path is $s \rightarrow b \rightarrow t$ with bottleneck capacity $\min(2,3) = 2$. Thus $v_G(\{a_2, a_3\}) = 2$.

\textbf{Coalition $C_3 = \{a_1, a_2\}$:} When only agents $a_1$ and $a_2$ cooperate (without $a_3$), the available edges are $E_{C_3} = \{e_1, e_2, e_3\}$. Critically, both edges leading to the sink ($e_4$ and $e_5$) are controlled by $a_3$ and are therefore unavailable. No flow can reach the sink, so $v_G(\{a_1, a_2\}) = 0$. This demonstrates that agent $a_3$ is essential for any positive flow.

\textbf{Grand Coalition $C_4 = \{a_1, a_2, a_3\}$:} When all agents cooperate, all edges are available. The maximum flow can be computed using standard algorithms (e.g., Ford-Fulkerson). Flow can be distributed as:
\begin{itemize}
    \item 2 units along $s \rightarrow a \rightarrow t$
    \item 1 unit along $s \rightarrow a \rightarrow b \rightarrow t$
    \item 2 units along $s \rightarrow b \rightarrow t$
\end{itemize}
This yields a total maximum flow of 5 units. Thus $v_G(\{a_1, a_2, a_3\}) = 5$.

\subsection{Key Observations}
This example demonstrates features that challenge learning-based approximation: (1) agent $a_3$ exhibits veto power with $v_G(C) = 0$ for all $C \not\ni a_3$, creating discontinuities in the characteristic function; (2) the grand coalition exhibits superadditivity through complementary paths: $v_G(N) = 5 > 4$ from pairwise sums; (3) exact computation requires $2^m$ maximum flow evaluations—for $m=20$, over one million computations per graph. These properties motivate our GNN-based approach to learn structural patterns of agent influence directly from graph topology.

\section{Experimental details}
\label{appendix:protocol_and_architectures}
This appendix provides the necessary details to ensure the full reproducibility of our experiments, including training and evaluation protocol, hyperparameters and model architectures.

\subsection{padding protocol}
For cross-configuration evaluation, models were trained and evaluated on datasets with matching feature dimensions. When comparing performance across different values of $n$ (graph size) or $p$ (edge density), we used unpadded data where edge features have dimension of $m+1$ and labels have dimensions $m$. For cross-m (different agent count) comparisons specifically, we trained separate models on zero-padded datasets where all configurations were padded to $m=20$: edge features were padded from $\mathbb{R}^{m+1}$to $\mathbb{R}^{21}$ and labels from $\mathbb{R}^{m}$ to $\mathbb{R}^{20}$ by appending zeros. This approach enabled meaningful comparison of how well models generalise across different numbers of agents while maintaining architectural compatibility. 

\subsection{Model architectures}
All models were implemented using common design patterns for stability and performance, including InstanceNorm for feature normalisation, GELU as the primary activation function, and residual connections where applicable. The specifics of the three main architectures are detailed below.

\subsubsection{Graph Isomorphism Network with Edge features}
Our GINE-based architecture uses GINEConv layers \cite{hu2021opengraphbenchmarkdatasets}, which extend the Graph Isomorphism Network\cite{xu2019powerfulgraphneuralnetworks} to incorporate edge features. The model consists of 5 GINE layers with a hidden dimension of 256. Each layer uses an MLP with an expansion structure (256, 1024, 256) and GELU activation. Edge features are encoded to the hidden dimension via a linear layer and incorporated into the message passing mechanism.

The model employs a late-fusion strategy: the GNN stacks processes only topological features (node type embeddings), while degree features are kept separate. Node embeddings are processed through an initial InstanceNorm layer. Each GINE layer is followed by InstanceNorm and includes residual connections where dimensionally compatible (note: the first layer has no residual as dimensions change from embedding to hidden size). After message passing, the graph embedding (obtained via global mean pooling) is concatenated with pooled raw degree features. This fused representation is processed by a 3-layer MLP (256,512, 256, output) with dropout (0.5) to produce final predictions.
The GINE architecture demonstrated stable training across different scales, with learning rate $10^{-4}$ and weight decay $10^{-4}$. Training time scaled sublinearly with graph size (roughly x2 slower for n=100 vs. n=20). 

\subsubsection{Graph Attention Networks}
Our GAT-based architecture \cite{veličković2018graphattentionnetworks} employs GATv2Conv layers \cite{brody2022attentivegraphattentionnetworks} with 4 attention heads across 4 layers (hidden dimensions 256). The model employs early fusion of node type embeddings with encoded degree features, pre-activation residual connections, and a Jumping Knowledge layer\cite{xu2018representationlearninggraphsjumping} to aggregate multi-scale representations. Graph-level predictions are made via global mean pooling of the JK output, which is then concatenated with pooled raw degree features (late fusion) before being processed by a 3-layer MLP. 

The GAT architecture required careful tuning: it showed sensitivity to learning rate, dataset size (struggling on smaller datasets with a tendency to predict mean values), and weight initialisation to avoid convergence to trivial solutions. Final training used learning rate $10^{-4}$ and weight decay $10^{-4}$, with Xavier uniform initialisation. Training time scaled approximately $x2$ between $n=20$ and $n=100$

\subsubsection{Graph Convolution Network with Edge-features}
Our EdgeConv architecture is based on NNConv layers \cite{gilmer2017neuralmessagepassingquantum}, which use edge-conditioned filters for message passing. The model employs a 3-layer structure with progressive channel expansion: the first layer maps from node embedding dimension to 128 channels, the second expands to 256 channels, and the third projects back to 128 channels (hidden dimension 128).

Edge features are processed by a small 2-layer MLP (with bottleneck dimensions 32 and GELU activation) that generates edge-specific weight matrices for each NNConv layer. Each convolution is followed by InstanceNorm, GELU activation, and dropout (0.2). To reduce memory consumption during training, we apply gradient checkpoints \cite{chen2016trainingdeepnetssublinear} to each convolution block using PyTorch checkpoint function.

Following the late-fusion pattern, the GNN processes only node type embeddings while degree features remain separate. After global mean pooling of both the graph embedding and raw degree features, they are concatenated and passed through a 3-layer MLP (of sizes 128,256,128, and then output) with dropout (0.2) for final predictions.

Due to edge-conditioned weight generation, EdgeConv's parameter count scales with the number of edges in the graph. This results in significantly longer training times for larger graphs: training on $n=100$ graphs took approximately $x15-x20$ longer than on $n=20$ graphs, compared to only ~$x2$ for GINE and GAT. To accommodate memory constraints, batch sizes were reduced for larger graphs (batch size 8 for $n=100$, 32 for $n=50$). Despite this computational cost, the architecture trained stably with a learning rate $10^{-4}$ and weight decay $10^{-4}$.

\section{Detailed GAT Results}
\label{appendix:gat_full_results}
We provide a detailed analysis of the GAT architecture following the main text evaluation structure: cross-edge density, cross-graph size, and cross-agent count. As described in the main text, GAT exhibits the poorest generalisation performance among the three architectures tested.

\subsection{Cross edge probability evaluations}
\begin{figure}[h]
     \centering
     \includegraphics[width=0.95\textwidth]{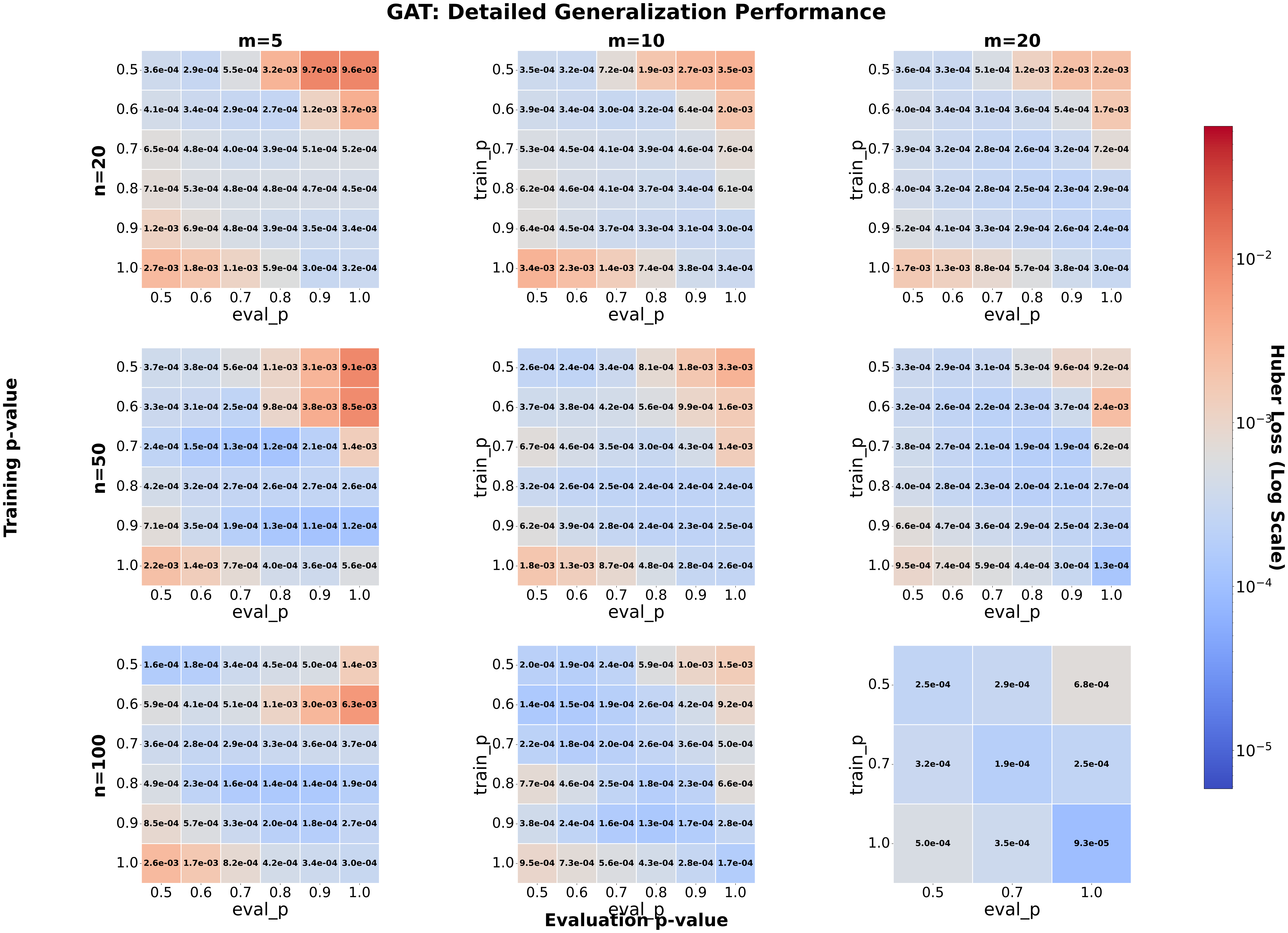}
     \caption{Detailed edge-probability generalisation analysis for GAT across all training configurations. Heatmaps show Huber loss (log scale) as a function of training edge probability (train\_p, y-axis) and evaluation edge probability (eval\_p, x-axis) for different graph sizes ($n=20, 50, 100$; rows) and agent counts ($m=5, 10, 20$; columns). GAT exhibits strongly asymmetric generalisation, with models trained on sparse graphs failing to transfer to dense configurations, and universal failure of $p=1.0$-trained models.}
     \label{fig:cross_p_gat}
\end{figure}

Figure \ref{fig:cross_p_gat} shows that GAT has substantial limitations in cross-edge probability generalisation. Unlike GINE and EdgeConv, GAT shows strongly asymmetric generalisation: models trained on lower edge probabilities (p<0.7) fail to generalise to higher densities, exhibiting significantly elevated loss values (often $x10-100$ times higher) compared to in-distribution performance. This asymmetry is unique to GAT, as the other architectures successfully generalise from sparse to dense graphs.

Consistent with all architectures, models trained at $p=1.0$ fail to generalise to sparser configurations across all graph sizes and agent counts. However, GAT's failure is particularly severe, with loss values reaching $10^-2$ or higher even for moderate sparsity differences. While $p=0.8$ provides slightly better bidirectional transfer, GAT still underperforms GINE and EdgeConv trained at any sparsity level. This suggests fundamental limitations in how attention mechanisms capture flow dynamics across varying edge densities.

\subsection{Cross graph size evaluations}
\begin{figure}[h]
     \centering
     \includegraphics[width=0.95\textwidth]{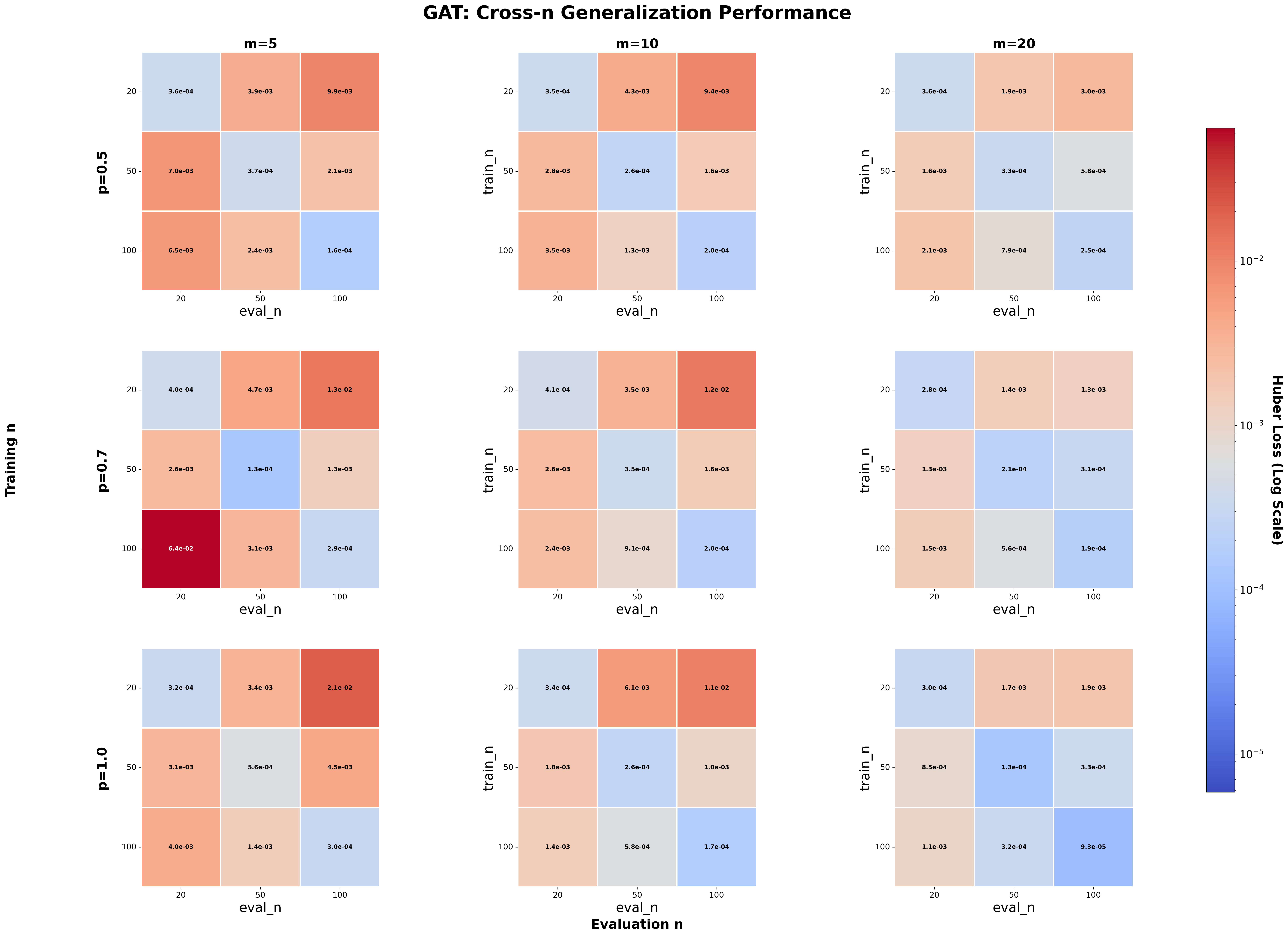}
     \caption{Detailed graph size generalisation analysis for GAT across all training configurations. Heatmaps show Huber loss (log scale) as a function of training number of nodes (train\_n, y-axis) and evaluation number of nodes (eval\_n, x-axis) for different edge probabilities ($p=0.5, 0.7, 1.$; rows) and agent counts ($m=5, 10, 20$; columns). Unlike GINE and EdgeConv, GAT exhibits poor generalisation in both directions, with particularly severe failure when trained on large graphs and evaluated on small ones. Diagonal elements consistently achieve the lowest error (in-distribution), and while best results for GAT, are still worse compared to in-distribution comparisons of GINE and EdgeConv.}
     \label{fig:cross_n_gat}
\end{figure}

Figure \ref{fig:cross_n_gat} demonstrates GAT's poor cross-size generalisation, with failure patterns in both directions, which is unique to this architecture. Models trained on small graphs ($n=20$) show moderate upward transfer to larger graphs, but with substantially higher error than GINE or EdgeConv. More critically, models trained on large graphs ($n=100$) catastrophically fail when evaluated on smaller graphs, particularly at $p=0.7$ where the train $n=100$, eval $n=20$ cell shows the Huber loss exceeding $10^-2$ (red colouring).

The bidirectional failure is particularly pronounced at intermediate sparsity ($p=0.7$), suggesting that GAT's attention mechanism struggles to learn scale-invariant representations of flow dynamics. At $p=1.0$, while GAT maintains relatively better cross-size transfer (consistent with other architectures), the absolute performance remains worse than GINE. The consistent poor performance across configurations indicates that GAT's attention-based aggregation may be overfitting to graph-size-specific patterns rather than learning generalisable influence structures.

\section{Detailed EdgeConv Results}
\label{appendix:edgeconv_full_results}

We provide a detailed analysis of the EdgeConv architecture following the main text evaluation structure. As described in the main text, EdgeConv achieved good generalisation results, second only to GINE, but suffers from substantial computational overhead due to parameter scaling with graph size, resulting in $x10-20$ longer training times and memory constraints for larger graphs.

\subsection{Cross edge probability evaluations}
\begin{figure}[h]
     \centering
     \includegraphics[width=0.95\textwidth]{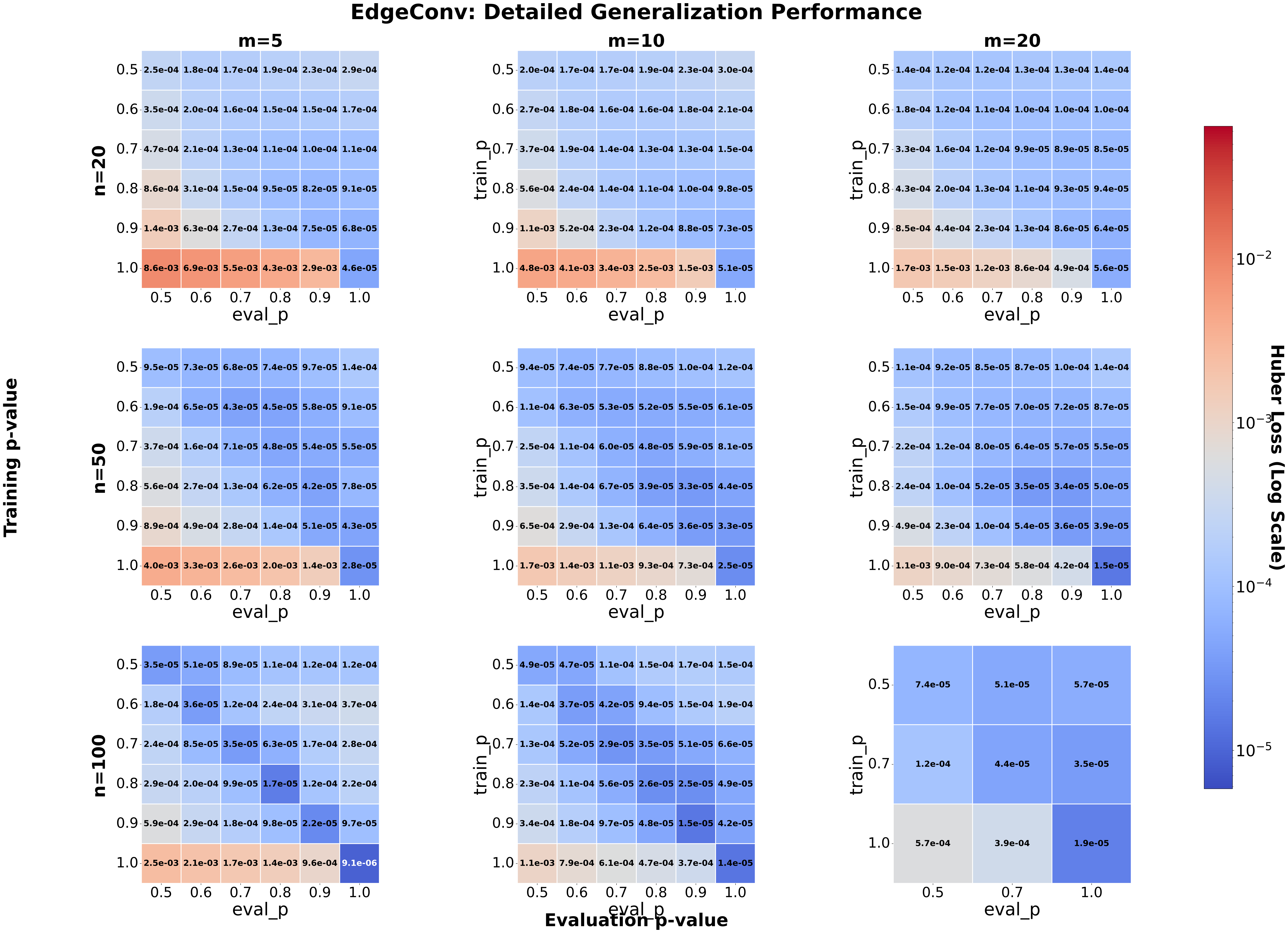}
     \caption{Detailed edge-probability generalisation analysis for EdgeConv across all training configurations. Heatmaps show Huber loss (log scale) as a function of training edge probability (train\_p, y-axis) and evaluation edge probability (eval\_p, x-axis) for different graph sizes ($n=20, 50, 100$; rows) and agent counts ($m=5, 10, 20$; columns). Diagonal elements consistently achieve the lowest error (in-distribution). EdgeConv demonstrates robust generalisation across edge densities, with optimal performance when trained at intermediate sparsity ($p\approx0.6$).}
     \label{fig:cross_p_edgeconv}
\end{figure}

Figure \ref{fig:cross_p_edgeconv} shows that EdgeConv achieves robust cross-edge probability generalisation comparable to GINE. Models trained at intermediate sparsity values ($p\approx0.6-0.7$) maintain low error across most evaluation densities, demonstrating effective learning density-invariant flow patterns. Unlike GAT, EdgeConv successfully generalises from sparse to dense graphs with minimal performance degradation.

The $p=1.0$ training limitation remains present: models trained on fully connected graphs fail to generalise to sparser configurations, consistent with all architectures. However, EdgeConv shows slightly more degradation in this regime compared to GINE, with Huber loss values approximately $x1.5-2$ higher when attempting to generalise from $p=1.0$ to $p<0.7$. The heatmaps reveal that training at $p=0.6$ provides optimal generalisation across the full density spectrum, suggesting that moderate sparsity encourages learning of transferable edge importance patterns. EdgeConv achieves this through edge-conditioned neural network transformations at each layer, offering an alternative mechanism to GINE's simpler aggregation scheme, though with greater computational overhead.

\subsection{Cross graph size evaluations}
\begin{figure}[h]
     \centering
     \includegraphics[width=0.95\textwidth]{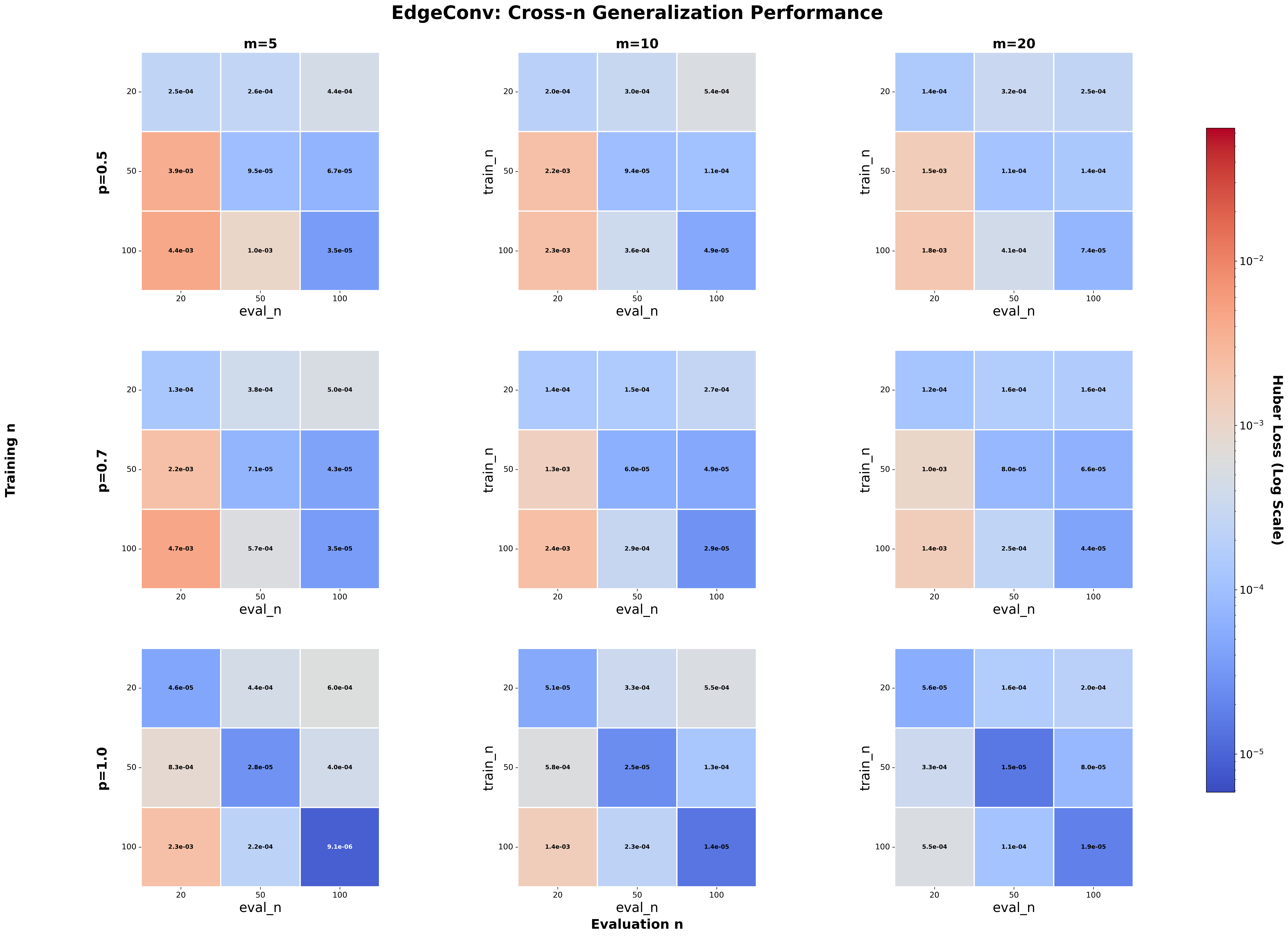}
     \caption{Detailed graph size generalisation analysis for EdgeConv across all training configurations. Heatmaps show Huber loss (log scale) as a function of training number of nodes (train\_n, y-axis) and evaluation number of nodes (eval\_n, x-axis) for different edge probabilities ($p=0.5, 0.7, 1.$; rows) and agent counts ($m=5, 10, 20$; columns). Diagonal elements consistently achieve the lowest error (in-distribution). EdgeConv demonstrates strong upward generalisation from small to large graphs, with optimal cross-size transfer at $p=1.0$.}
     \label{fig:cross_n_edgeconv}
\end{figure}

Figure \ref{fig:cross_n_edgeconv} reveals EdgeConv's strong directional generalisation pattern, closely matching GINE's behaviour. Models trained on small graphs ($n=20$) successfully transfer upwards to larger graphs ($n=50,100$) with minimal error increase, while models trained on large graphs show moderate degradation when evaluated on smaller graphs. This asymmetry is less severe than GAT's bidirectional failure but slightly more pronounced than GINE's.

Edgeconv exhibits particularly strong cross-size transfer at $p=1.0$, where models trained at any graph size maintain relatively uniform low error across all evaluation sizes, consistent with the pattern observed for GINE. This suggests that fully-connected topologies may provide training signals that transfer more readily across scales, though the mechanism remains unclear. At lower densities ($p=0.5,0.7$), EdgeConv maintains the directional generalisation pattern described in the main text: upward transfer from smaller to larger graphs succeeds, while downward transfer shows more moderate degradation. Despite EdgeConv's computational inefficiency for large graphs, its generalisation quality makes it a viable alternative to GINE when training on smaller representative graphs for evaluation across multiple scales.

\section{Detailed computational analysis}
\label{appendix:comp_analysis}

Table \ref{table:computational_comparison} presents detailed timing comparisons across all architectures and representative configurations. We report three metrics: training time per graph (amortized over 200K training instances), inference time per graph, and total time (sum of amortized training and inference costs).

Several patterns emerge. First, Monte Carlo's computational cost scales with both agent count and graph size. Increasing agents from$ m=5$ to $m=20$ (rows 2 to 4) increases inference time from 2.92ms to 475ms—a $x163$ slowdown reflecting the $\mathcal{O}(2^m)$ complexity of coalition enumeration. Graph size also impacts performance substantially: at $m=20$, $p=1.0$, increasing from $n=20$ to $n=50$ to $n=100$ increases inference time from 475ms to 3.84s to 20.07s, reflecting the $\mathcal{O}(E \cdot V^2)$ maximum flow computation cost. In contrast, GNN inference time remains nearly constant (0.6-0.7ms) across all configurations, demonstrating the method's scalability advantages.

Second, the amortised training cost becomes negligible for repeated evaluations. For the largest configuration ($n=100$,$ m=20$, $p=1.0$), GINE requires 60ms amortised training time per graph but only 0.62ms inference time. The breakeven point occurs after just two graph evaluations: total cost per graph (30.3ms) already outperforms Monte Carlo (20.07s per graph), a $x660$ speedup. This advantage compounds with scale: evaluating 1000 graphs costs 620 seconds with GINE versus 20,070 seconds with Monte Carlo.

Third, architectural differences in training cost become pronounced for larger graphs. EdgeConv requires 528ms per graph for $n=100$ ($x8.8$ slower than GINE), while GAT requires 109ms ($x1.8$ slower). However, all GNN architectures maintain sub-millisecond inference regardless of training cost.

The cross-configuration example (rows 7-8) demonstrates zero-shot generalization benefits: a model trained on $ n=50$ graphs evaluates$ n=100$ graphs with only inference cost (0.70ms), whereas training directly on $n=100$ incurs 60ms amortised cost: a $x85$ reduction in total time while maintaining prediction accuracy (as shown in Figure \ref{fig:cross_n_gine}).
  
\begin{table*}[t]
\caption{Computational performance comparison. Time in seconds per graph. Training time amortised over 200K graphs. MC: Monte Carlo($N=10,000$), EC: EdgeConv.}
\label{table:computational_comparison}
\centering
\setlength{\tabcolsep}{1.0pt}
\rotatebox{90}{
\begin{tabular}{c|c|cccc|cccc|cccc|lll}
\multirow{2}{*}{\shortstack{Training \\ dataset}} & \multirow{2}{*}{\shortstack{Evaluation \\ dataset}} & \multicolumn{4}{c|}{\shortstack{Training time \\ per graph (sec)}}                                                                             & \multicolumn{4}{c|}{\shortstack{Inference time \\ per graph (sec)}}                                                                            & \multicolumn{4}{c|}{\shortstack{Total time \\ per graph (sec)}}                                                                               &  &  &  \\ \cline{3-14}
                                  &                                     & \multicolumn{1}{l|}{MC} & \multicolumn{1}{l|}{GINE}    & \multicolumn{1}{l|}{GAT}     & \multicolumn{1}{l|}{EC} & \multicolumn{1}{l|}{MC} & \multicolumn{1}{l|}{GINE}    & \multicolumn{1}{l|}{GAT}     & \multicolumn{1}{l|}{EC} & \multicolumn{1}{l|}{MC} & \multicolumn{1}{l|}{GINE}    & \multicolumn{1}{l|}{GAT}     & \multicolumn{1}{l|}{EC} &  &  &  \\ \cline{1-14}
n=20,m=5,p=0.5                    & n=20,m=5,p=0.5                      & \multicolumn{1}{c|}{2.1e-3}      & \multicolumn{1}{c|}{4.37e-2} & \multicolumn{1}{c|}{4.03e-2} & 4.68e-2                       & \multicolumn{1}{c|}{2.1e-3}      & \multicolumn{1}{c|}{5.68e-4} & \multicolumn{1}{c|}{6.06e-4} & 7.24e-4                       & \multicolumn{1}{c|}{4.21e-3}     & \multicolumn{1}{c|}{4.43e-2} & \multicolumn{1}{c|}{4.09e-2} & 4.75e-2                       &  &  &  \\
n=20,m=5,p=1.0                    & n=20,m=5,p=1.0                      & \multicolumn{1}{c|}{2.92e-3}     & \multicolumn{1}{c|}{4.35e-2} & \multicolumn{1}{c|}{3.21e-2} & 3.93e-2                       & \multicolumn{1}{c|}{2.92e-3}     & \multicolumn{1}{c|}{5.89e-4} & \multicolumn{1}{c|}{6.42e-4} & 8.19e-4                       & \multicolumn{1}{c|}{5.84e-3}     & \multicolumn{1}{c|}{4.41e-2} & \multicolumn{1}{c|}{3.27e-2} & 4.01e-2                       &  &  &  \\
n=20,m=10,p=1.0                   & n=20,m=10,p=1.0                     & \multicolumn{1}{c|}{1.87e-1}     & \multicolumn{1}{c|}{4.35e-2} & \multicolumn{1}{c|}{3.21e-2} & 4.39e-2                       & \multicolumn{1}{c|}{1.87e-1}     & \multicolumn{1}{c|}{6.05e-4} & \multicolumn{1}{c|}{6.59e-4} & 8.23e-4                       & \multicolumn{1}{c|}{3.74e-1}     & \multicolumn{1}{c|}{4.40e-2} & \multicolumn{1}{c|}{3.27e-2} & 4.47e-2                       &  &  &  \\
n=20,m=20,p=1.0                   & n=20,m=20,p=1.0                     & \multicolumn{1}{c|}{4.75e-1}     & \multicolumn{1}{c|}{4.56e-2} & \multicolumn{1}{c|}{4.04e-2} & 6.61e-2                       & \multicolumn{1}{c|}{4.75e-1}     & \multicolumn{1}{c|}{5.97e-4} & \multicolumn{1}{c|}{6.41e-4} & 7.47e-4                       & \multicolumn{1}{c|}{9.51e-1}     & \multicolumn{1}{c|}{4.62e-2} & \multicolumn{1}{c|}{4.10e-2} & 4.11e-2                       &  &  &  \\
n=50,m=20,p=1.0                   & n=50,m=20,p=1.0                     & \multicolumn{1}{c|}{3.84}        & \multicolumn{1}{c|}{3.79e-2} & \multicolumn{1}{c|}{6.83e-2} & 3.31e-1                       & \multicolumn{1}{c|}{3.84}        & \multicolumn{1}{c|}{6.84e-4} & \multicolumn{1}{c|}{7.27e-4} & 1.75e-3                       & \multicolumn{1}{c|}{7.68}        & \multicolumn{1}{c|}{3.86e-2} & \multicolumn{1}{c|}{6.90e-2} & 3.32e-1                       &  &  &  \\
n=100,m=20,p=1.0                  & n=100,m=20,p=1.0                    & \multicolumn{1}{c|}{20.07}       & \multicolumn{1}{c|}{5.98e-2} & \multicolumn{1}{c|}{1.09e-1} & 5.28e-1                       & \multicolumn{1}{c|}{20.07}       & \multicolumn{1}{c|}{6.20e-4} & \multicolumn{1}{c|}{7.68e-4} & 2.51e-3                       & \multicolumn{1}{c|}{40.17}       & \multicolumn{1}{c|}{6.04e-2} & \multicolumn{1}{c|}{1.09e-1} & 5.30e-1                       &  &  &  \\
n=50,m=20,p=1.0                   & n=100,m=20,p=1.0                    & \multicolumn{1}{c|}{3.84}        & \multicolumn{1}{c|}{3.79e-2} & \multicolumn{1}{c|}{6.83e-2} & 3.31e-1                       & \multicolumn{1}{c|}{20.07}       & \multicolumn{1}{c|}{7.03e-4} & \multicolumn{1}{c|}{7.26e-4} & 2.49e-3                       & \multicolumn{1}{c|}{23.91}       & \multicolumn{1}{c|}{3.86e-2} & \multicolumn{1}{c|}{6.90e-2} & 3.33e-1                       &  &  &  \\
n=20,m=5,p=0.5                    & n=20,m=5,p=1.0                      & \multicolumn{1}{c|}{2.1e-3}      & \multicolumn{1}{c|}{4.37e-2} & \multicolumn{1}{c|}{4.03e-2} & 4.68e-2                       & \multicolumn{1}{c|}{1.87e-1}     & \multicolumn{1}{c|}{6.04e-4} & \multicolumn{1}{c|}{6.06e-4} & 8.08e-4                       & \multicolumn{1}{c|}{1.89e-1}     & \multicolumn{1}{c|}{4.43e-2} & \multicolumn{1}{c|}{4.09e-2} & 4.76e-2                       &  &  & 
\end{tabular}
}
\end{table*}

\end{document}